\documentclass[10pt,twocolumn,letterpaper]{article}

\usepackage{cvpr}              

\usepackage[accsupp]{axessibility}
\usepackage{graphicx}
\usepackage{graphics} 
\usepackage{amsmath}
\usepackage{amssymb}
\usepackage{booktabs}
\usepackage{float}
\usepackage{enumitem}
\usepackage[dvipsnames]{xcolor}
\usepackage{booktabs}
\usepackage{float}
\usepackage{url}
\usepackage{color, colortbl}
\usepackage{cellspace}
\usepackage[pagebackref,breaklinks,colorlinks]{hyperref}

\renewcommand{\theenumi}{\Alph{enumi}}

\definecolor{Gray}{gray}{0.9}
\definecolor{LightCyan}{rgb}{0.88,1,1}
\definecolor{sh_gray}{rgb}{0.84,0.84,0.84}
\definecolor{sh_gray2}{rgb}{1,0.89,0.75}
\definecolor{color3}{rgb}{0.95,0.95,0.95}
\definecolor{color4}{rgb}{0.96,0.96,0.86}
\definecolor{color5}{rgb}{0.90,0.90,0.90}

\usepackage[capitalize]{cleveref}
\crefname{section}{Sec.}{Secs.}
\Crefname{section}{Section}{Sections}
\Crefname{table}{Table}{Tables}
\crefname{table}{Tab.}{Tabs.}

\def\cvprPaperID{5} 
\def\confName{CVPR}
\def\confYear{2022}

\begin{document}

\title{\LARGE \bf
Exploring Map-based Features for Efficient Attention-based\\Vehicle Motion Prediction
}

\author{Carlos Gómez-Huélamo$^{1}$, Marcos~V. Conde$^{2}$, Miguel Ortiz$^{1}$\\
$^{1}$Department of Electronics, University of Alcal{\'a} (UAH), Spain.\\
$^{2}$Computer Vision Lab, Institute of Computer Science, University of Würzburg, Germany\\
{\tt\small {carlos.gomezh@uah.es}, {marcos.conde-osorio@uni-wuerzburg.de}}\\
{\small Authors contributed equally to this work.}
}

\maketitle

\begin{abstract}
Motion prediction (MP) of multiple agents is a crucial task in arbitrarily complex environments, from social robots to self-driving cars. Current approaches tackle this problem using end-to-end networks, where the input data is usually a rendered top-view of the scene and the past trajectories of all the agents; leveraging this information is a must to obtain optimal performance. In that sense, a reliable Autonomous Driving (AD) system must produce reasonable predictions on time, however, despite many of these approaches use simple ConvNets and LSTMs, models might not be efficient enough for real-time applications when using both sources of information (map and trajectory history). Moreover, the performance of these models highly depends on the amount of training data, which can be expensive (particularly the annotated HD maps). In this work, we explore how to achieve competitive performance on the Argoverse 1.0 Benchmark using efficient attention-based models, which take as input the past trajectories and map-based features from minimal map information to ensure efficient and reliable MP. These features represent interpretable information as the driveable area and plausible goal points, in opposition to black-box CNN-based methods for map processing.
Our code is publicly available at \href{https://github.com/Cram3r95/mapfe4mp }{https://github.com/Cram3r95/mapfe4mp}.
\end{abstract}


\section{Introduction}
\label{sec:introduction}


\begin{figure}[!ht]
  \centering
   \includegraphics[width=0.83\linewidth]{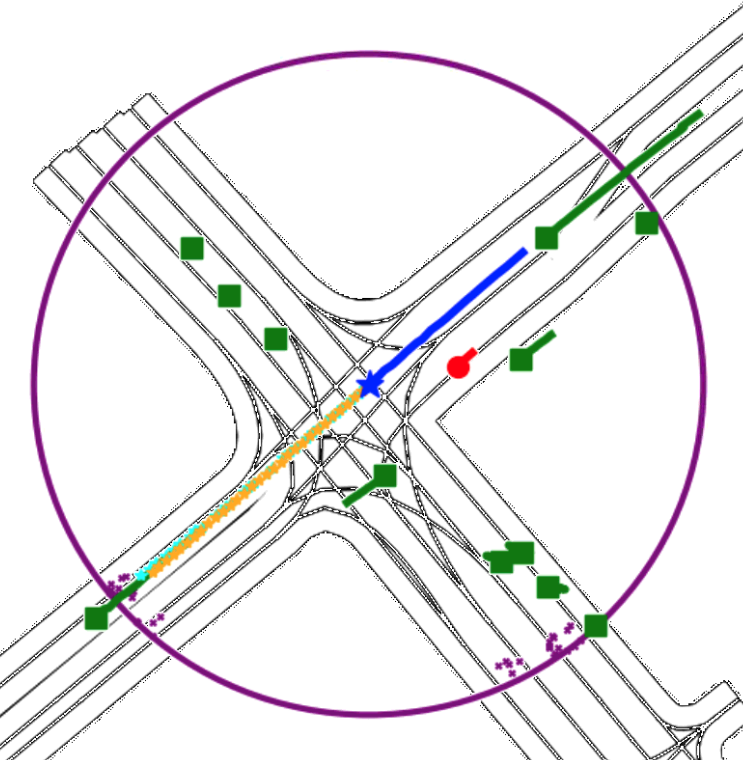}
   \caption{Motion Prediction Scenario in Argoverse~\cite{chang2019argoverse}. We represent: our vehicle (\textcolor{red}{ego}), the target \textcolor{blue}{agent}, and \textcolor{ForestGreen}{other agents}. We can also see the \textcolor{cyan}{real} trajectory, the \textcolor{orange}{prediction}, the estimated \textcolor{purple}{action range} and potential \textcolor{purple}{goal-points}. Markers are current positions.}
   \label{fig:teaser}
\end{figure}


Autonomous Driving (AD) is one of the most challenging research topics in academia and industry due to its real-world impact: improved safety, reduced congestion and greater mobility~\cite{KATRAKAZAS2015416realtime, vitelli2021safetynet, filos2020av, Geiger2012urban}. Predicting the future behavior of traffic agents around the ego-vehicle is one of the key unsolved challenges in reaching full self-driving autonomy~\cite{kesten2019lyft, houston2020onelyftdata}. In that sense, an AD stack can be hierarchically broken down in the following tasks: (i) perception, identify what is around us, (ii) track and predict, what will happen next, (iii) planning and decision-making, deciding what the AD stack is going to do in the future, (iv) control, send the corresponding commands to the vehicle. Assuming the surrounding agents have been detected and tracked (we have their past trajectories), the final task of the perception layer is known as Motion Prediction (MP), that is, predict the future trajectories ~\cite{mahjourian2022occupancy, trajectron, weng2021mtptraject, Ivanovic_2019_ICCVtrajectron} of surrounding traffic agents given the past sensor information and considering the corresponding traffic rules and social interaction among the agents, which is usually task full of uncertainty. 

These predictions are required to be multi-modal, which means given the past motion of a vehicle and the surrounding scene, there may exist more than one possible future behaviour (also known as modes), therefore models need to cover the different choices a driver could make (i.e. going straight or turning, accelerations or slowing down) as a possible trajectory in the immediate future or as a probability distribution of the agent's future location~\cite{gilles2021home, dendorfer2020goalgan, dong2021multimodal}.

Traditional methods for motion forecasting~\cite{hong2019rules, KATRAKAZAS2015416realtime, gomez2021smartmot} are based on physical kinematic constraints and road map information with handcrafted rules. Though these approaches are sufficient in many simple situations (i.e. cars moving in constant velocity), they fail to capture the rich behavior strategies and interaction in complex scenarios.

On the other hand, Deep Learning (DL) advances allow us to understand and capture the complexity of a driving scenario using data-driven methods~\cite{deo2018convolutionalmotion, casas2018intentnet, chai2019multipath, bock2019ind, can2022maps, Xu_2017_CVPR} and achieve the most promising \textit{state-of-the-art} (SOTA) results by learning such intrinsic rules.
These models need to take into account the local High Definition (HD) Map~\cite{can2022maps}, the past trajectory of the participant agents and the interactions with other actors, where obtaining and fusing this information is a research topic by itself~\cite{trajectron, varadarajan2021multipath++, Prakash_2021_CVPR} and a core part in the AD pipeline. In that sense, researchers have identified a bottleneck for efficient real-time applications~\cite{KATRAKAZAS2015416realtime, Romera2018erfnet, Rella2021decocermp}, as usually, more data-inputs implies higher complexity and inference time~\cite{gao2020vectornet}.\\

In this paper, following the same principles as recent SOTA methods, we aim to achieve competitive results that ensure reliable predictions, as observed in Fig.~\ref{fig:teaser}, yet, using light-weight attention-based models that take as input the past trajectories of each agent, and integrate prior-knowledge about the map easily. In that sense, we made the following contributions: (1) Identify a key problem in the size of motion prediction models, with implications in real-time inference and edge-device deployment, (2) Propose a motion prediction model that does not explicitly rely on HD context maps (either vectorized or rasterized), which can be useful as HD context maps can be costly to obtain, (3) Use fewer operations (FLOPs) than other SOTA models to achieve competitive performance on Argoverse 1.0~\cite{chang2019argoverse, argobench}.

\section{Related work}
\label{sec:related_work}

One of the crucial tasks that Autonomous Vehicles (AV) must face during navigation, specially in arbitrarily complex urban scenarios, is to predict the behaviour of dynamic obstacles \cite{lerner2007ucydata, pellegrini2009ethdata, chang2019argoverse, bock2019ind, trajectron, Scheel2022urbanreal}. In a similar way to humans that pay more attention to close obstacles, people walking towards them or upcoming turns rather than considering the presence of building or people far away, the perception layer of a self-driving car must be modelled to focus more on the salient regions of the scene \cite{sadeghian2019sophie, dendorfer2020goalgan} and the more relevant agents to predict the future behaviour of each traffic participant. In that sense, high-fidelity maps~\cite{can2022maps} have been widely adopted to provide offline information (also known as physical context) to complement the online information provided by the sensor suite of the vehicle and its corresponding algorithms. Recent learning-based approaches \cite{mahjourian2022occupancy, xiao2022multimodalend2end, casas2018intentnet, deo2018cstlstmpool, rhinehart2019precog, ivanovic2021heterogeneous}, which present the benefit of having probabilistic interpretations of different behaviour hypotheses, require to build a representation to encode the trajectory and map information. Hong et al.~\cite{hong2019rules} assumes that detections around the vehicle are provided and focuses on behaviour prediction by encoding entity interactions with ConvNets. \textbf{Intentnet}~\cite{casas2018intentnet} proposes to jointly detect traffic participants (mostly focused on vehicles) and predict their trajectories using raw LiDAR pointcloud and HD map information. 
\textbf{MultiPath} by Chai et al.~\cite{chai2019multipath} uses ConvNets as encoder and adopts pre-defined trajectory anchors to regress multiple possible future trajectories.
\textbf{SoPhie} by Sadeghian et al.~\cite{sadeghian2019sophie} is an interpretable trajectory prediction framework based on Generative Adversarial Network (GAN) and attention mechanisms, which leverages two sources of information: the path history of all the agents in a scene, and the scene context information, using images of the scene. This model can successfully generate multiple physically acceptable paths that respect social constraints of the environment. Authors show that by modeling jointly the information about the physical environment and interactions between all agents, the model is able to learn better than when both sources of information are used independently.

\begin{figure}[!h]
    \centering
    \setlength{\tabcolsep}{2.0pt}
    \begin{tabular}{c}
    \includegraphics[width=\linewidth]{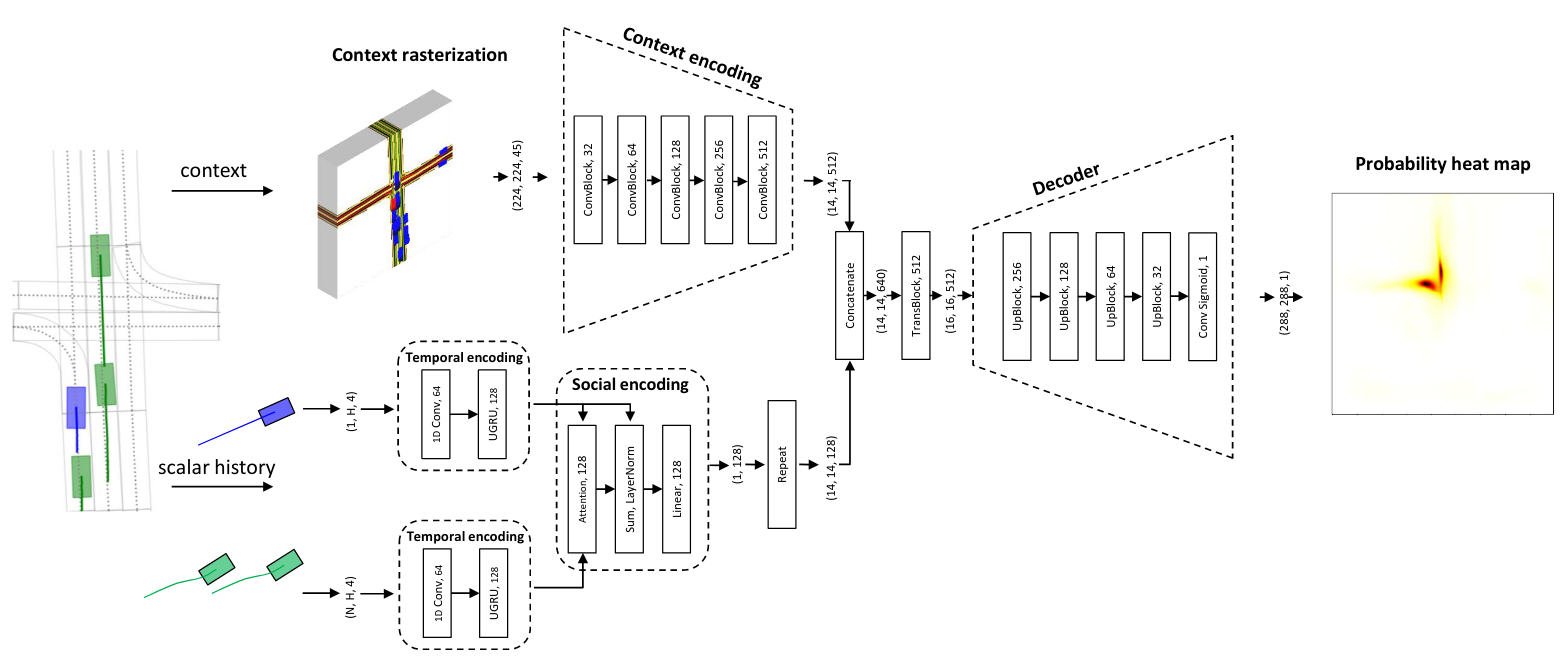} \tabularnewline
    HOME~\cite{gilles2021home} \tabularnewline
    \tabularnewline
    \includegraphics[width=\linewidth]{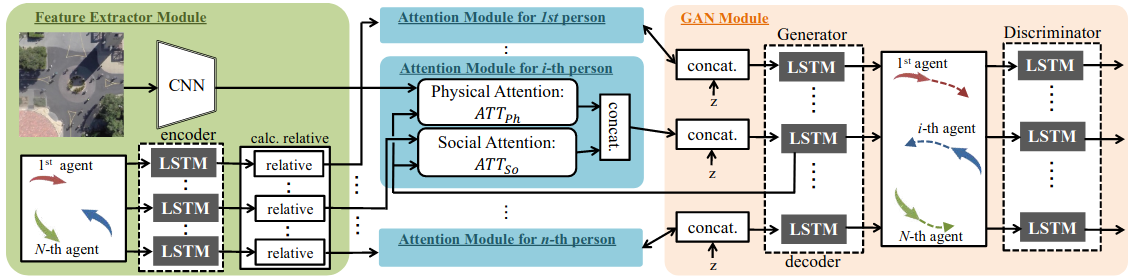} \tabularnewline
    SoPhie~\cite{sadeghian2019sophie} \tabularnewline
    \end{tabular}
    \caption{We show the architectures of two \textit{state-of-the-art} frameworks for MP. These methods leverage both the physical context (map), previous behaviour (past trajectories) and interaction among agents (social encoding) to generate realistic trajectories.}
    \label{fig:prev_models}
\end{figure}

\textbf{HOME} by Gilles et al.~\cite{gilles2021home} presented a novel representation for multimodal trajectory prediction, where the model takes as input the context (map) and history of past trajectories, and generates an unconstrained 2D heatmap representation of the agent’s possible future trajectories, which represents the probability distribution of the agent’s future location. The method builds on simple architecture with classic convolution networks coupled with attention mechanism for agent interactions. Fig.~\ref{fig:prev_models} shows the SoPhie and HOME architectures. Both methods have two-streams, one for each input (trajectories and map), and use CNNs to process the map or physical context.

Other approaches based on Graph Neural Networks (GNNs) such as \textbf{GOHOME}~\cite{gilles2021gohome} and \textbf{LaneGCN}~\cite{liang2020learninggraph} have achieved SOTA results on the most relevant benchmarks for Motion Prediction~\cite{gilles2021gohome, liang2020learninggraph, chandra2020forecastinggraphlstm}. Moreover, despite GAN-based approaches~\cite{sadeghian2019sophie, julka2021conditionalgan, dendorfer2020goalgan, gupta2018sgan} provide certain control and interpretability, most competitive approaches on self-driving motion prediction benchmarks such as Argoverse~\cite{chang2019argoverse} or Waymo~\cite{Sun_2020_CVPR_waymodata} do not use adversarial training, where the training complexity is one the main reasons.

As observed, DL methods are data-hungry and use complex features (both social and physical) to predict, in an accurate way, the future behaviour of the agents in the scene. Most \textit{state-of-the-art} MP methods require an overwhelmed amount of information as input, specially in terms of the physical context, such as the above mentioned HD maps. This might be inefficient in terms of latency, parallelism or computational cost (MACs, FLOPs)~\cite{gao2020vectornet, walters2020trajectory}. In Section~\ref{sec:ours} we explain our solution to this trade-off problem.

\subsection{Datasets}
\label{sec:datasets}

Large-scale annotated datasets have been proposed to impulse the research in this field.
ETH~\cite{pellegrini2009ethdata} and UCY~\cite{lerner2007ucydata} datasets contain annotated trajectories of real world pedestrians interacting in a variety of real social situations.
Focusing on self-driving cars, we find NuScenes Prediction~\cite{caesar2020nuscenes}, the Waymo Motion Prediction~\cite{gu2021densetntwaymo} dataset and the Lyft MP~\cite{houston2020onelyftdata, kesten2019lyft} dataset, which contains over 1000 hours of data, providing the physical context usually as a birds-eye-view (BEV) representation of the scene and few seconds of past trajectories for each agent in the scene.

Due to our particular interest in agents with non-holonomic constraints (that is, the agent cannot conduct an extremely abrupt change of motion between consecutive timestamps), in this work we focus on the \textbf{Argoverse} Motion Forecasting Dataset~\cite{chang2019argoverse}. This dataset was collected by a fleet of autonomous vehicles, and includes more than 300,000 5-second tracked scenarios with a particular vehicle identified for trajectory forecasting.

\begin{figure*}[!ht]
    \centering
    \setlength{\tabcolsep}{2.0pt}
    \begin{tabular}{cc}
    \includegraphics[width=0.65\linewidth]{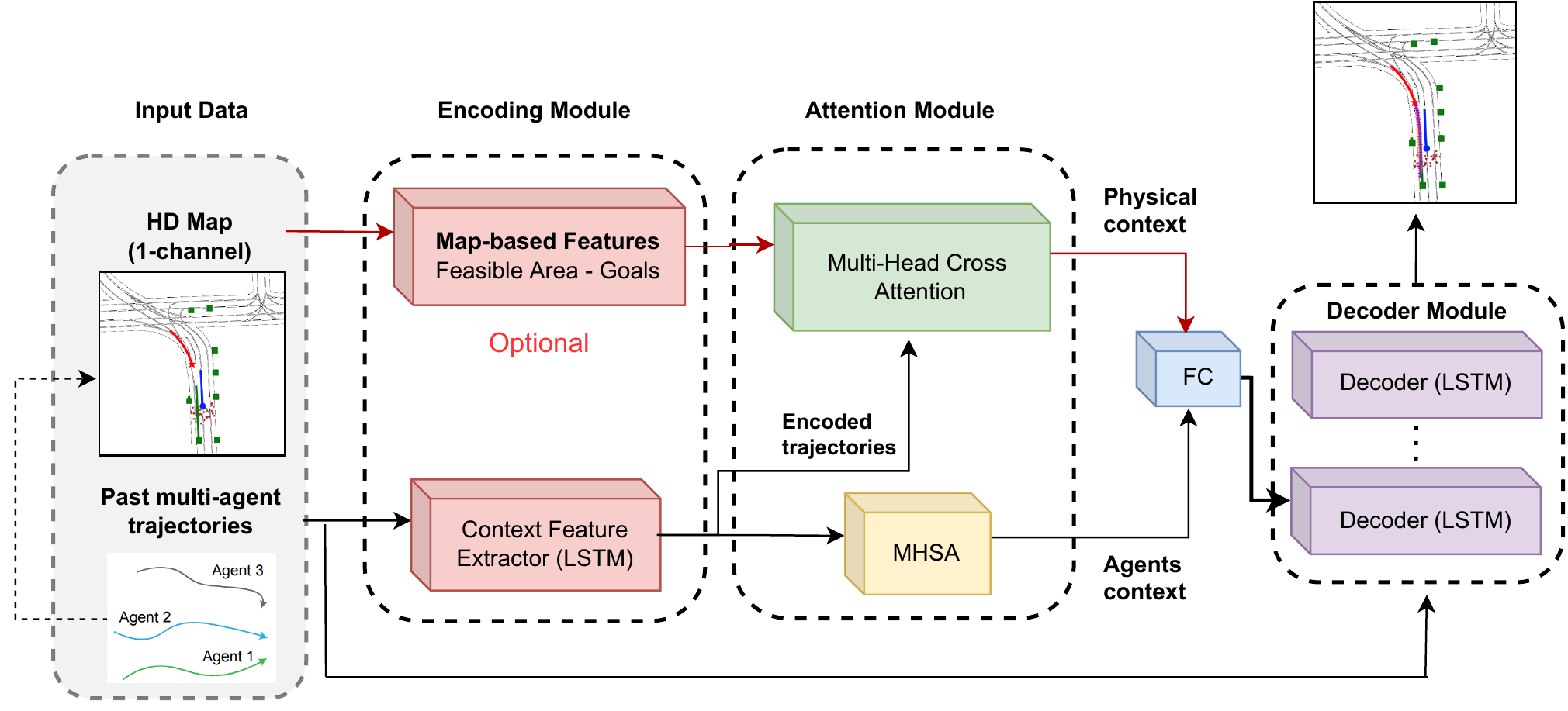} &
    \includegraphics[width=0.35\linewidth]{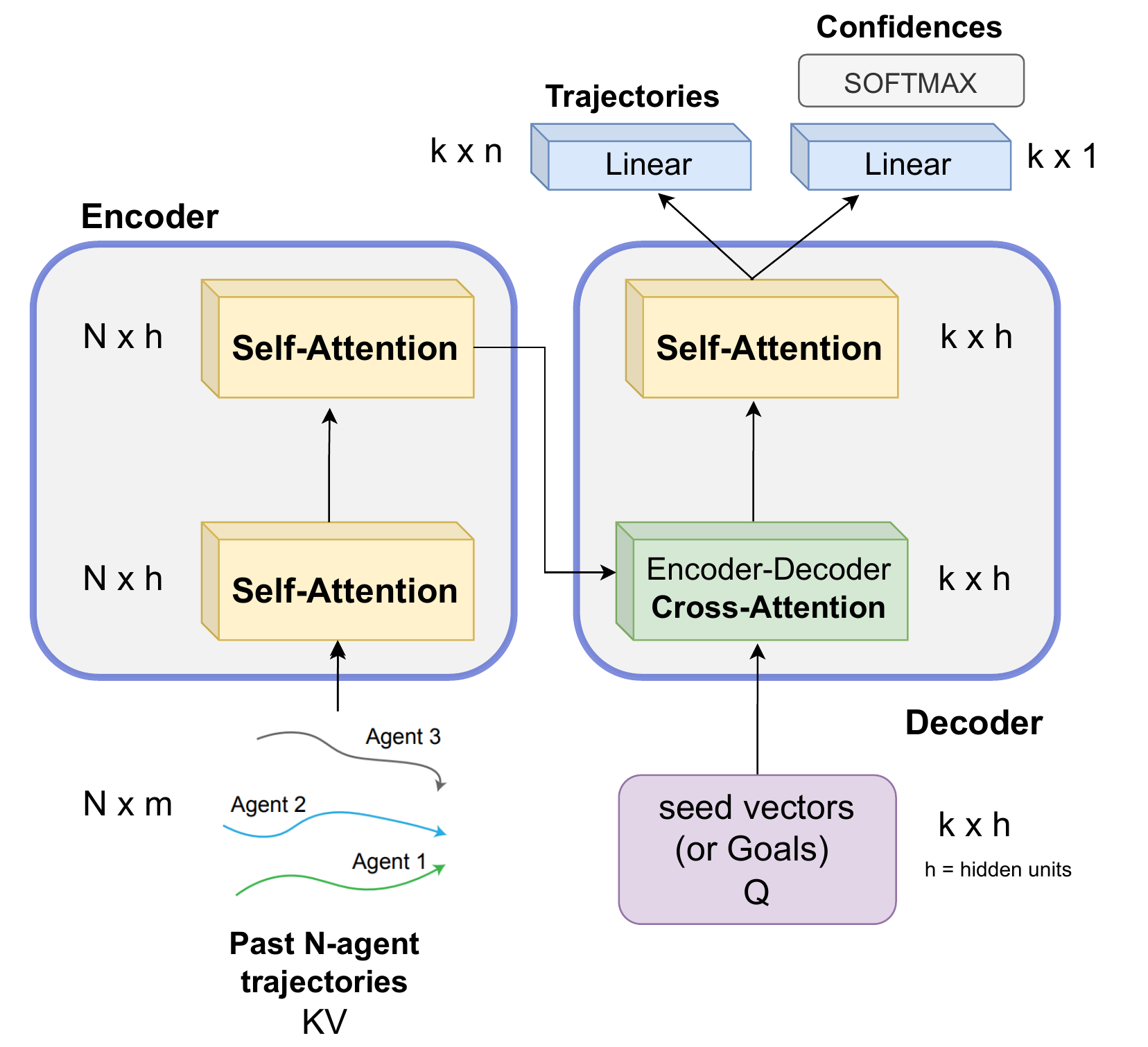}
    \tabularnewline
    LSTM MHSA~\cite{lstm, zhang2019multi, mercat2020multiattentmotion} & Set Transformer~\cite{lee2019set, girgis2021autobots} \tabularnewline
    \end{tabular}
    \caption{Overview of our attention-based MP Model. We distinguish three main blocks: 1) Target points extractor, which uses HDMap information and agent past trajectory to compute plausible target (destination) points, 2) Multi-Head Self-Attention (MHSA) for encoding the trajectories of the surrounding vehicles and learning context, 3) Decoder and trajectory generator. We show the two versions of our model using (a) LSTM~\cite{lstm} and (b) Set Transformer~\cite{lee2019set}.}
\label{fig:main_diagram}
\end{figure*}

\section{Our approach}
\label{sec:ours}

Considering the trade-off between curated input data and complexity, we aim to achieve competitive results on the well-known Argoverse Benchmark~\cite{chang2019argoverse, argobench} using a single data source (past trajectories) and a simplified representation of the feasible area around the vehicle. Therefore, our models do not require full-annotated HD Maps or BEV representations of the scene in order to compute the physical context. We use a traditional motion forecasting in order to exploit this simplified representation of the feasible area and compute goal points as prior information for further improvement performance by feeding this information into the model about the physically feasible areas where the agents can interact. 

Moreover, we use traditional motion forecasting to exploit the availability of simple maps, and further improve performance by feeding information into the model about the physically feasible areas where the agents can interact. Fig.~\ref{fig:main_diagram} shows an overview of our approach.
\subsection{Experimental Setup}
\label{sec:experiments}

\paragraph{Dataset}
We use the public available Argoverse Motion Forecasting Dataset~\cite{chang2019argoverse} which consists of 205942 training samples, 39472 validation samples and 78143 test samples. Data was sampled at 10 Hz, where each sample contains the position (x|y) of all agents in the scene in the past 2s (20 observed points), the local map, and the labels are the 3s (30 predicted points) future positions of one target agent in the scene. 

\paragraph{Metrics}

We evaluate the performance of our models using the following standard metrics~\cite{chang2019argoverse, Sun_2020_CVPR_waymodata}: (i) Final Displacement Error (FDE), which computes the mean of the $l_2$ distance between the final points of the ground truth and the predicted final position. (ii) Average Displacement Error (ADE), which averages the distances between the ground truth and predicted output across all timesteps.
When the output is multimodal, we generate K outputs (also known as modes) per prediction step and report the metrics for the best out of the K outputs (i.e. minFDE$_k$). We will report results for $K=1$ and $K=6$ as this is the standard in the Argoverse Motion Forecasting dataset in order to compare with other models.

\subsection{Models}

As stated before, our models take as input the history of $m=20$ past observations (2s) for each agent in the scene, defining its base trajectory. The models output, in the uni-modal case, is a trajectory of $n=30$ points (3s) for the target agent; in the multi-modal case the output are $k$ trajectories of $n$ points. This can be seen as a \textit{seq2seq}~\cite{sun20183dofseq, park2018sequence, zheng2021rethinkingseq} problem.

\subsubsection{LSTM based with Multi-Head Self-Attention}
\label{sec:lstm_model}

Aiming at efficiency, we design a Multi-Head Self-Attention (MHSA) \textbf{LSTM}~\cite{lstm, mercat2020multiattentmotion, zhang2019multi}. Similar to the methods proposed by Mercat et al.~\cite{mercat2020multiattentmotion} and Sadeghian et al.~\cite{sadeghian2019sophie} we propose a LSTM encoder-decoder module with a self-attention mechanism that allows agent interactions (input trajectories) while being invariant to their number and ordering. In this context, each vehicle should pay attention to specific features from a selection of the other vehicles.
The encoder takes as input the past trajectories from each agent in the scene, and applies $b$ self-attention blocks to them to produce feature vectors that encode all pairwise relations between agents' information. The MHSA consists of 4 heads, generating the agents context from the encoded trajectories. The decoder uses the context from MHSA to predict $n$ future points (see Fig.~\ref{fig:main_diagram}).


\subsubsection{Set transformer}
\label{sec:transformer_model}

\textbf{Transformers}~\cite{vaswani2017attention} provide the most powerful attention mechanism, in exchange, model's efficiency is compromised as the complexity grows. As an alternative, we use Set Transformer~\cite{lee2019set}, which is invariant to the permutation of the input trajectories and does not utilize neither positional encoding nor layer normalization or residual connections, and therefore, it is less complex than classical Transformers~\cite{lee2019set, sanakoyeu2021lyft}. We show Set Transformer architecture in Fig.~\ref{fig:main_diagram}. The encoder applies $b$ self-attention blocks to them to produce feature vectors that encode all pairwise relations between the input agents trajectories. The first block of the decoder is a cross-attention block, which takes $k$ learnable (query) seed vectors (Q) for our $k$ output trajectories, and the encoded representation of the input trajectories (KV). Next follows $s$ self-attention blocks after which we produce $k$ output trajectories and the corresponding confidences. Alternative, we can use prior-knowledge as Q instead of the learnable vectors, as we introduce next in Sec.~\ref{subsec:map-based}.

As mentioned in Section~\ref{sec:related_work}, this module (attention) is fundamental in SOTA frameworks for motion prediction (i.e. HOME~\cite{gilles2021home} or SoPhie~\cite{sadeghian2019sophie}). We can further improve this by adding additional information about each agent such as velocity and orientation, or information about the scene.\\

\subsubsection{Map-based Features} 
\label{subsec:map-based}
Multiple approaches have tried to predict realistic trajectories by means of learning physically feasible areas as heatmaps or probability distributions of the agent’s future location~\cite{dendorfer2020goalgan, sadeghian2019sophie, gilles2021home}. These approaches require either a top-view RGB BEV image of the scene, or a HD Map with exhaustive topological, geometric and semantic information (commonly codified as channels). This information is usually encoded using a CNN and fed into the model together with the social agent's information~\cite{dendorfer2020goalgan, sadeghian2019sophie, gao2020vectornet}.

\begin{figure}[!ht]
    \centering
    \setlength{\tabcolsep}{2.0pt}
    \begin{tabular}{ccc}
    \includegraphics[width=0.30\linewidth]{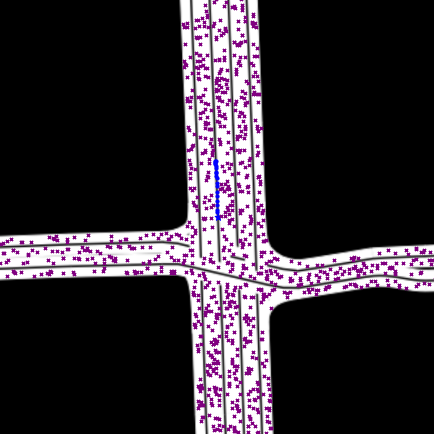} & 
    \includegraphics[width=0.30\linewidth]{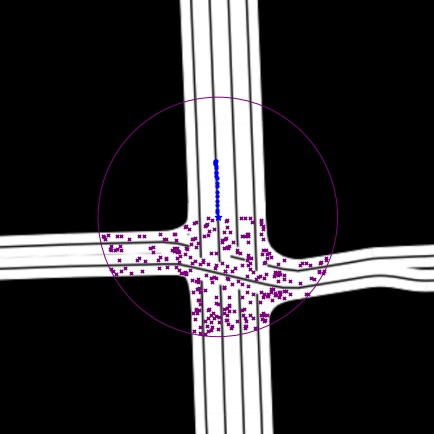}
    \includegraphics[width=0.30\linewidth]{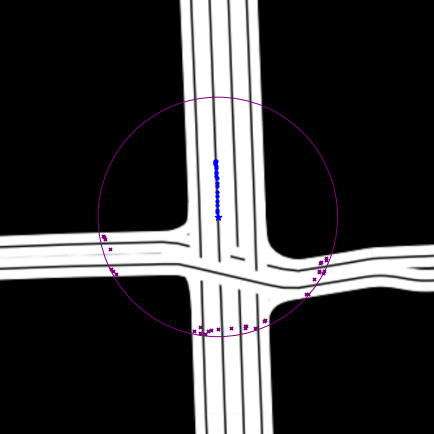} 
    \end{tabular}
    \caption{Goal Points Estimation from the Feasible area. Left: Random sampling in the discrete Feasible Area around the target agent. Dynamic filter (considering orientation and velocity in the last observation frame). Right: Final proposals}
    \label{fig:goal-points}
\end{figure}

In this work we estimate the range of motion (360º) using a minimal HD Map representation that includes only the feasible area (white area in Fig.~\ref{fig:goal-points}), where we can discretize the feasible area $\mathcal{F}$ (represented by a discrete grid of the \textit{width} x \textit{height} BEV map image where the pixels are driveable) as a subset of $r$ randomly sampled points $\{p_0 , p_1 ... p_r\}$ from such area in the map (easy to extract from a 1-channel binarized HD image) considering the orientation and velocity in the last observation frame for the agent. This step can be considered as pre-processing of the HD Map, therefore the model never sees the HD map image nor the whole graph of nodes. 
Fig.~\ref{fig:goal-points} summarizes step-by-step the whole process. First, we calculate the driveable area (white area in Fig.~\ref{fig:goal-points}) around the vehicle considering a hand-defined \textit{d} threshold.

Then, we consider the dynamic features of the target agent in the last observation frame $t_{obs}$ to compute acceptable target points in local coordinates. Nevertheless, in Argoverse benchmark 1.1, these dynamic features are not provided, in such a way they must be calculated. Since the trajectory data are noisy with tracking errors, as expected from a real-world dataset, simply interpolating the coordinates between consecutive time steps, assuming constant frequency, results in noisy estimation. Then, in order to estimate the orientation and velocity of the target agent in the last observation frame $t_{obs}$, we compute a vector for each feature given $\theta_{i}=\arctan{(\frac{y_{i}-y_{i-1}}{x_{i}-x_{i-1}})}$ and $v_{i}=\frac{X_{i}-X_{i-1}}{t_{i}-t_{i-1}}$, where $X_{i}$ represents the 2D position of the agent at each observed frame $i$ as state above. Once both vectors are computed, we obtain a smooth estimation as proposed by \cite{tang2021exploring} of the heading angle (orientation) and velocity by assigning less importance (higher forgetting factor) to the first observations observations, in such a way immediate observations are the key states to determine the current spatio-temporal variables of the agent, as in the following equation: 

\begin{equation}
    \hat{\psi}_{T_h} = \sum_{t=0}^{T_h}\lambda^{T_h - t}\psi_t
    \label{eq:dynamic_feats_last_observation_frame}
\end{equation}

where $T_h$ is the number of observed frames, $\psi_t$ is the estimated orientation/velocity at the $t_{i}$ frame, $\lambda\in(0, 1)$ is the forgetting factor, and $\hat{\psi}_{t_obs}$ is the smoothed orientation/velocity estimation at the last observed frame. After estimating these variables, we calculate the range of motion around the target agent as a circle with radius: $H * \psi_{v}$ where $\psi_{v}$ is the estimated velocity using Eq.~\ref{eq:dynamic_feats_last_observation_frame} and $H=3$ is the time-horizon of 3s. Finally we randomly sample $r$ points $p \in \mathcal{F}$ in this range considering a constant velocity model during the prediction horizon and the estimated orientation, assuming non-holonomic constraints \cite{triggs1993motion} which are inherent of standard road vehicles, that is, the car has three degrees of freedom, its position in two axes and its orientation, and must follow a smooth trajectory in a short-mid term prediction.

This representation not only reunites information about the feasible area around the agent, but also represents potential \textbf{Goal points}~\cite{dendorfer2020goalgan} (i.e. potential destinations or end-of-trajectory points for the agents). Moreover, this information is \textit{"cheap"} and \textit{interpretable}, therefore, we do not need further exhaustive annotations from the HD Map in comparison with other methods like HOME, which gets as input a 45-channel encoded map~\cite{gilles2021home}.
We concatenate this information, as 2D vector $\mathcal{V}$, together with the model's social context features to generate more realistic trajectories (see Fig.~\ref{fig:main_diagram}). 






\subsection{Implementation Details}

We train our models to convergence using a single NVIDIA RTX 3090, and validate our results on the official Argoverse validation set~\cite{argobench}. We use Adam optimizer with learning rate $0.001$ and default parameters, batch size $64$ and linear LR Scheduler with factor $0.5$ decay on plateaus.\\
\textbf{Augmentations:} (i) Dropout random points from the past trajectory, (ii) random $90^{\circ}$ rotations of the trajectories, (iii) point location perturbations under a $\mathcal{N}(0, 0.2)$ [m] distribution~\cite{ye2021tpcn}.\\

\textbf{Loss:} We calculate the negative log-likelihood (NLL) of the ground truth points $\mathbf{g}=\{(x_0,y_0) ... (x_n, y_n)\}$ given the $k$ modalities (hypotheses) $\mathbf{p}=\{(\hat{x}^1_0,\hat{y}^1_0) ... (\hat{x}^k_n, \hat{y}^k_n)\}$, with confidences $\mathbf{c}=\{c_1 ... c_k\}$ using the following equation:

\begin{equation}
    \text{NLL} = -\log \sum_{k} e^{ \log{c^k} - \frac{1}{2} \sum_{t=0}^n (\hat{x}^k_t - x_t)^2 + (\hat{y}^k_t - y_t )^2 }
\label{eq:nll}
\end{equation}

Similar to ~\cite{mercat2020multiattentmotion}, we assume the ground truth positions to be modeled by a mixture of multi-dimensional independent Normal distributions over time (predictions with unit covariance). We also calculate the following error indicators:

\begin{equation}
  \label{eq:ade}
  \begin{gathered}
    \text{ADE} = \dfrac{1}{n} \sum_{t=0}^{n} \sqrt{(x_t-\hat{x}_t)^2 + (y_t-\hat{y}_t )^2}
  \end{gathered}
\end{equation}

\begin{equation}
  \label{eq:fde}
  \begin{gathered}
    \text{FDE} = \lVert \mathbf{g}_n - \mathbf{p}_n \rVert_{2}^{2}
  \end{gathered}
\end{equation}

Minimizing the NLL loss maximizes the likelihood of the data for the forecast, however, it tends to overfit part of its output~\cite{mercat2020multiattentmotion}. For this reason we also add as regularization the ADE and FDE (introduced in Sec.~\ref{sec:experiments}). Therefore, our loss function is:

\begin{equation}
    \mathcal{L} = \alpha \text{NLL} + \beta \text{ADE} + \gamma \text{FDE}
\label{eq:loss}
\end{equation}

Where $\alpha=0.75$ , $\beta=1$ and $\gamma=0.5$ initially, and can be manually adjusted during training (especially $\gamma$).

\subsection{Results}
\label{sec:results}

\begin{table*}
  \centering
  \caption{Results on the Argoverse Benchmark~\cite{chang2019argoverse}. We borrow some numbers from~\cite{chang2019argoverse, gilles2021home, gilles2021gohome}. We specify the inputs for each model: HD Map or vectorized map, and Trajectories. Our models used the simplified binarized HD Map* with only the feasible area annotated. Our best model has $0.1$M parameters in comparison with HOME~\cite{gilles2021home} $5$M, yet, the difference in ADE is only $0.02$.}
  \begin{tabular}{lc>{\columncolor[gray]{0.9}}c>{\columncolor[gray]{0.9}}c c c}
    \toprule
    \rowcolor[gray]{0.9} Model & Input & \multicolumn{2}{c}{K=1} & \multicolumn{2}{c}{K=6}\\
    & & ADE~(m) $\downarrow$ & FDE~(m) $\downarrow$ & minADE~(m) $\downarrow$ & minFDE~(m) $\downarrow$ \\
    \midrule
    Constant Velocity~\cite{chang2019argoverse} & & 3.53 & 7.89 &  &  \\ 
    Argoverse Baseline (NN)~\cite{chang2019argoverse} & & 3.45 & 7.88 & 1.71 & 3.29 \\
    SGAN~\cite{gupta2018sgan} & Map + Traj. & 3.61 & 5.39 &  &  \\
    TPNet~\cite{fang2020tpnet} & Map + Traj. & 2.33 & 5.29 &  &   \\
    Challenge Winner: uulm-mrm (2nd)~\cite{chang2019argoverse} & Map + Traj. & 1.90 & 4.19 & 0.94 & 1.55 \\
    Challenge Winner: Jean (1st)~\cite{mercat2020multiattentmotion, chang2019argoverse} & Map + Traj. & 1.74 & 4.24 & 0.98 & 1.42 \\
    TNT~\cite{zhao2020tnt} & Map + Traj. & 1.77 & 3.91 & 0.94 & 1.54 \\
    mmTransformer~\cite{liu2021mmtransf} & Map + Traj. & 1.77 & 4.00 & 0.84 &  1.33 \\
    HOME~\cite{gilles2021home} & Map + Traj. & 1.72 & 3.73 & 0.92 & 1.36 \\
    LaneConv~\cite{deo2018convolutionalmotion} & Map + Traj. & 1.71 & 3.78 & 0.87 & 1.36 \\
    UberATG~\cite{liang2020learninggraph} & Map + Traj. & 1.70 & 3.77 & 0.87 & 1.36 \\
    LaneRCNN~\cite{zeng2021lanercnn} & Map + Traj. & 1.70 & 3.70 & 0.90 & 1.45 \\
    GOHOME~\cite{gilles2021gohome} & Map + Traj. & 1.69 & 3.65 & 0.94 & 1.45 \\
    TPCN~\cite{ye2021tpcn} & Map + Traj. & 1.58 & 3.49 & 0.82 & 1.24 \\
    \textbf{State-of-the-art (top-10)}~\cite{gilles2021gohome, liu2021mmtransf, varadarajan2021multipath++, ye2021tpcn} & Map + Traj. & \textbf{1.57}$\pm$0.06 &  \textbf{3.44}$\pm$0.15 & \textbf{0.79}$\pm$0.02 & \textbf{1.17}$\pm$0.04  \\
    \textbf{State-of-the-art (top-25)}~\cite{gilles2021gohome, liu2021mmtransf, varadarajan2021multipath++, ye2021tpcn} & Map + Traj. & \textbf{1.63}$\pm$0.08 & \textbf{3.59}$\pm$0.20 & \textbf{0.81}$\pm$0.03 & \textbf{1.22}$\pm$0.06  \\
    \midrule
    Ours LSTM (Unimodal) & Traj. & 1.80 & 4.02 &  & \\
    Ours Transformer & Traj. & 1.77 & 3.90 & 0.94 & 1.45   \\
    Ours LSTM + Map Features (Goals) & Map* + Traj. & 1.76 & 3.91 &  & \\
    Ours Transformer + Map Features (Goals)  & Map* + Traj. & 1.74 & 3.84 & 0.92 & 1.40 \\
    \bottomrule
  \end{tabular}
  \label{tab:ablation}
\end{table*}

\begin{table}[!ht]
    \centering
    \caption{Efficiency comparison among SOTA methods. We show the number of parameters for each model, ADE, and FLOPs.
    }
    \resizebox{\linewidth}{!}{
    \begin{tabular}{lccccc}
         Model & \# Params.~(M) & ADE~(m) $\downarrow$ & FLOPs (G)~$\downarrow$ \\
         \toprule
         HOME~\cite{gilles2021home} & 5.1 & 1.72 & 4.80 \\
         GOHOME~\cite{gilles2021gohome} & 0.40 & \textbf{1.69} & 0.09 \\
         CtsConv~\cite{walters2020trajectory} & 1.08 & 1.85 & 4.32 \\
         R18-k3-c1-r100~\cite{gao2020vectornet} & 0.25  & 2.21 & 0.66 \\
         R18-k3-c1-r400~\cite{gao2020vectornet} & 0.25 & 2.16 & 10.56 \\
         VectorNet~\cite{gao2020vectornet} & \textbf{0.07} & 1.81 & 0.041\\
         Ours & 0.1 & 1.74 & \textbf{0.018} \\
         \bottomrule
    \end{tabular}}
    \label{tab:effcomp}
\end{table}


As we state in Section~\ref{sec:introduction} and~\ref{sec:ours} our main goal is to achieve competitive results while being efficient in terms of model complexity; in particular in terms of FLOPs (Floating-Point Operations per second). For this reason, we have proposed light-weight models, whose main input is the history of past trajectories of the agents, complemented by interpretable map-based features (Goal points). In this section we analyze our results and ablation studies, and prove the possible benefits of our approach for self-driving motion prediction.

We aim to compare our performance with \textbf{VectorNet} by Gao et al.~\cite{gao2020vectornet}, which is a well-known efficient method, for this reason, we also focus on uni-modal predictions.

The Argoverse Benchmark~\cite{argobench} has over 200 submitted methods, however, the top approaches have, in our opinion, essentially the same performance. In order to do a fair comparison, we analyze the \textit{state-of-the-art} performance in this benchmark, we show the results in Table~\ref{tab:ablation}. Given the standard deviations (in meters), we can conclude that there is no significant performance difference for the top-25 models.

\begin{figure}[!ht]
    \centering
    \setlength{\tabcolsep}{2.0pt}
    \begin{tabular}{cc}
    \includegraphics[width=0.5\linewidth]{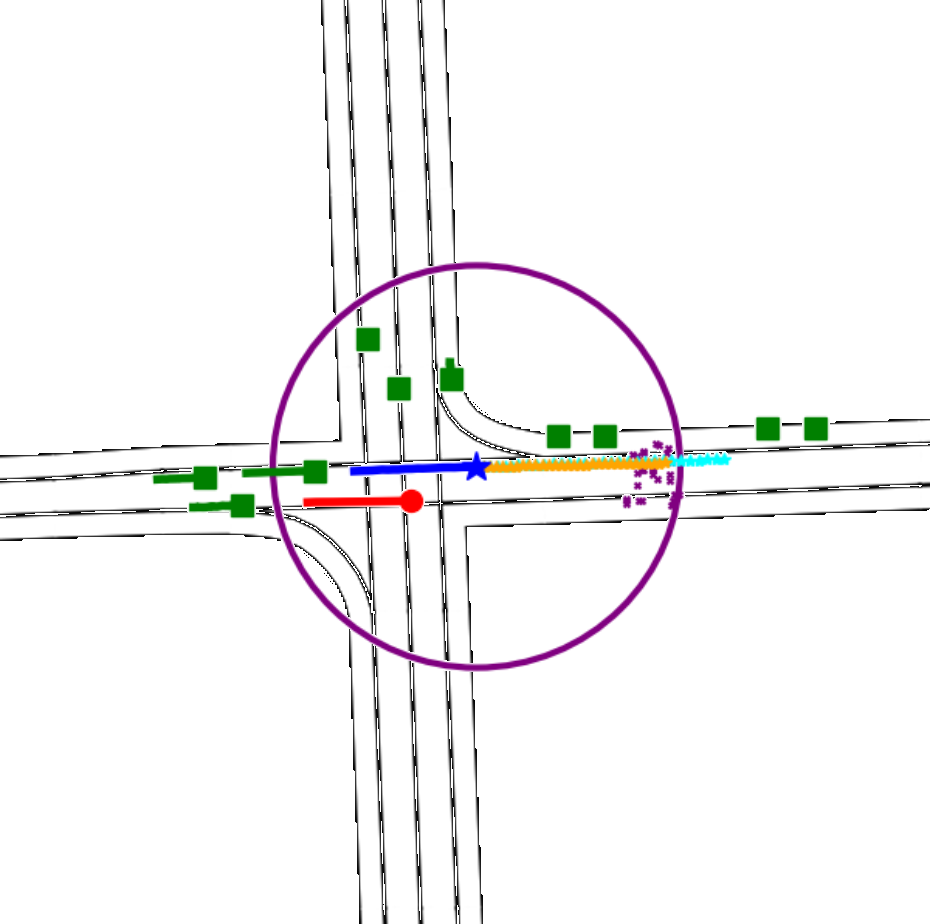} & 
    \includegraphics[width=0.5\linewidth]{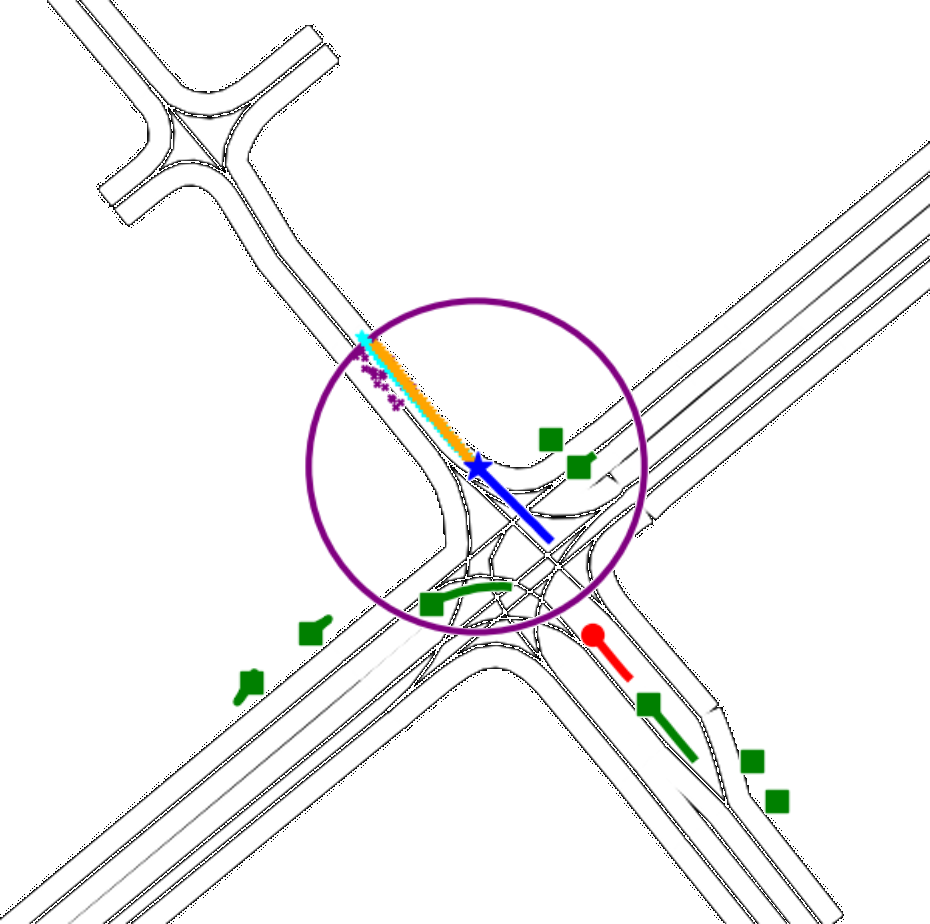}
    \tabularnewline
    \includegraphics[width=0.5\linewidth]{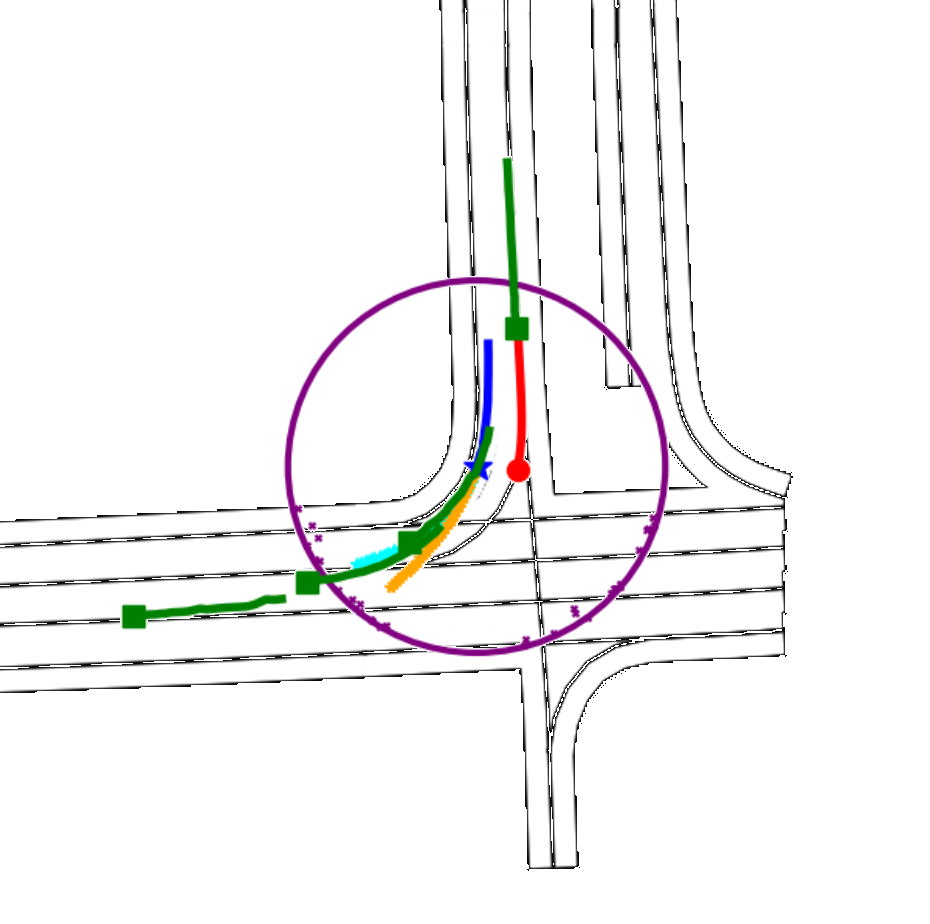} & 
    \includegraphics[width=0.5\linewidth]{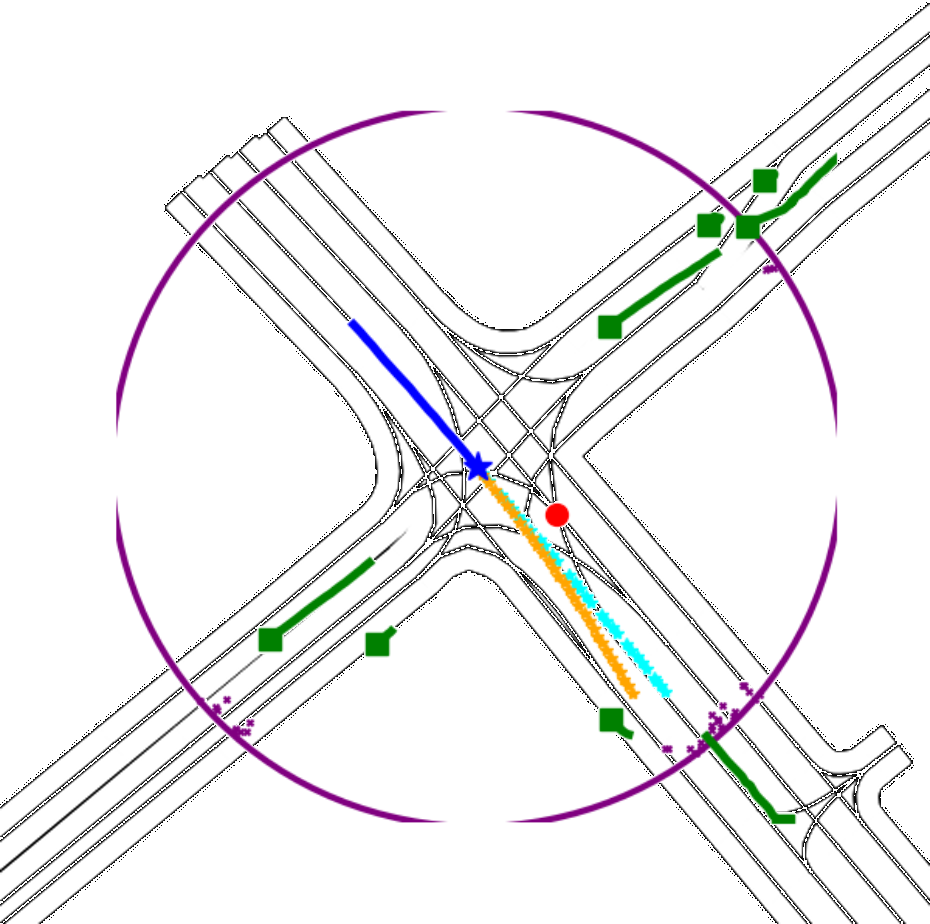}
    \tabularnewline
    \end{tabular}
    \caption{Uni-modal Qualitative Results. We show the target \textcolor{blue}{agent}, the \textcolor{cyan}{real} trajectory, and the \textcolor{orange}{prediction}. We use the same legend as in Fig.~\ref{fig:teaser}.}
    \label{fig:results_teaser}
\end{figure}
\begin{figure*}[!ht]
    \centering
    \setlength{\tabcolsep}{2.0pt}
    \begin{tabular}{cccc}
    \fbox{\includegraphics[width=0.23\linewidth]{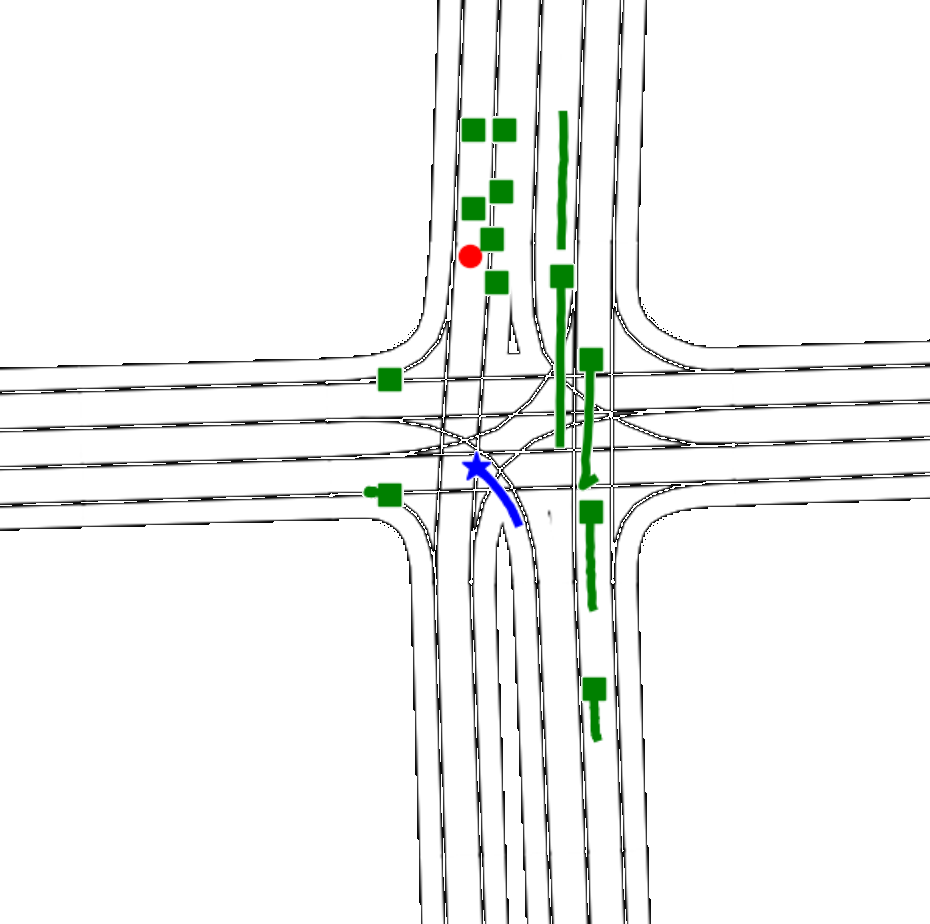}} & 
    \fbox{\includegraphics[width=0.23\linewidth]{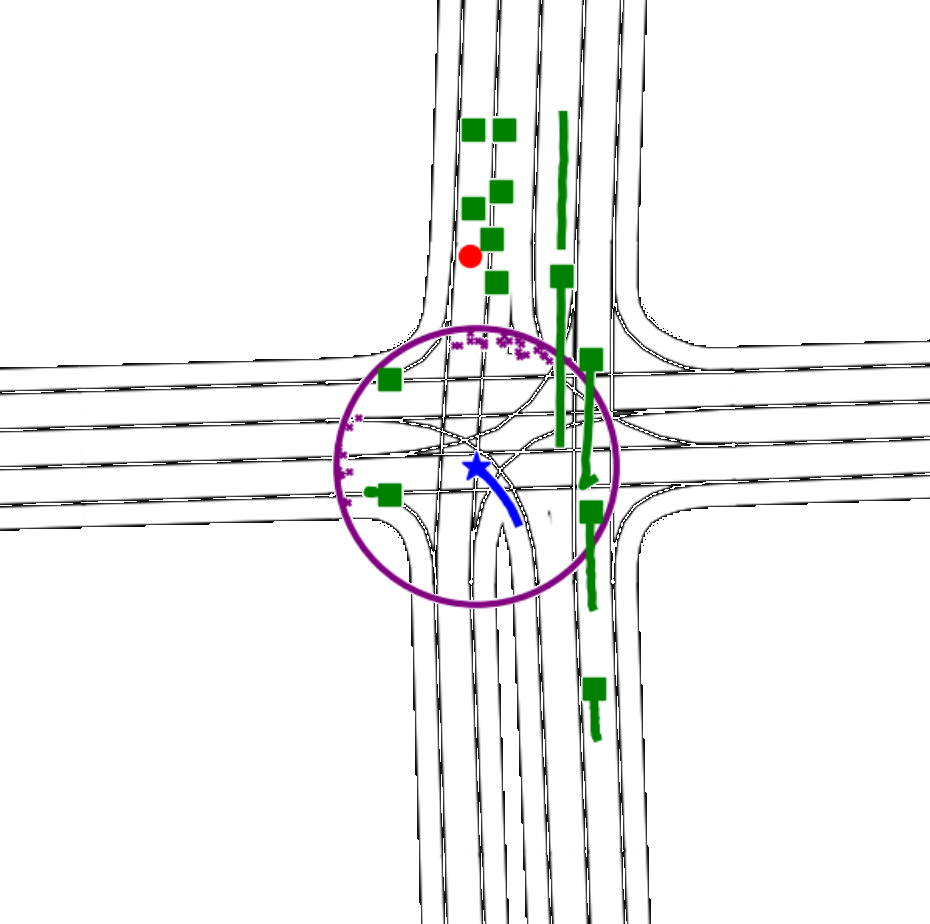}} &
    \fbox{\includegraphics[width=0.23\linewidth]{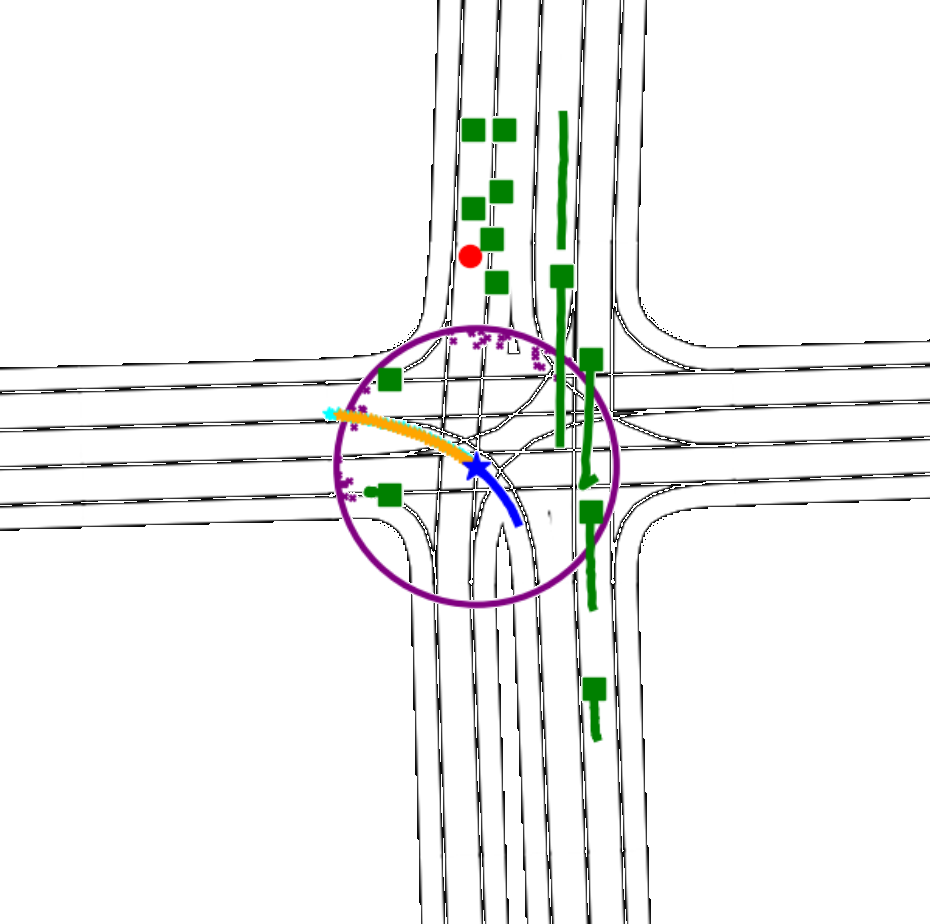}} & 
    \fbox{\includegraphics[width=0.23\linewidth]{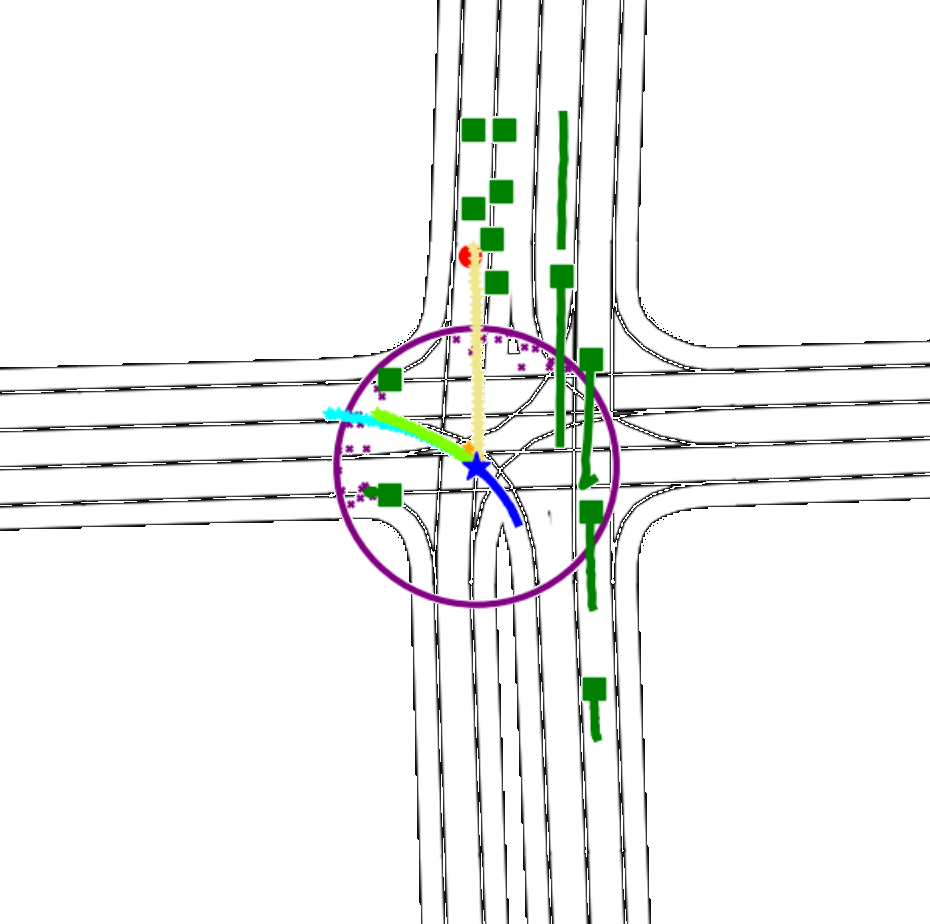}}
    \tabularnewline
    \fbox{\includegraphics[width=0.23\linewidth]{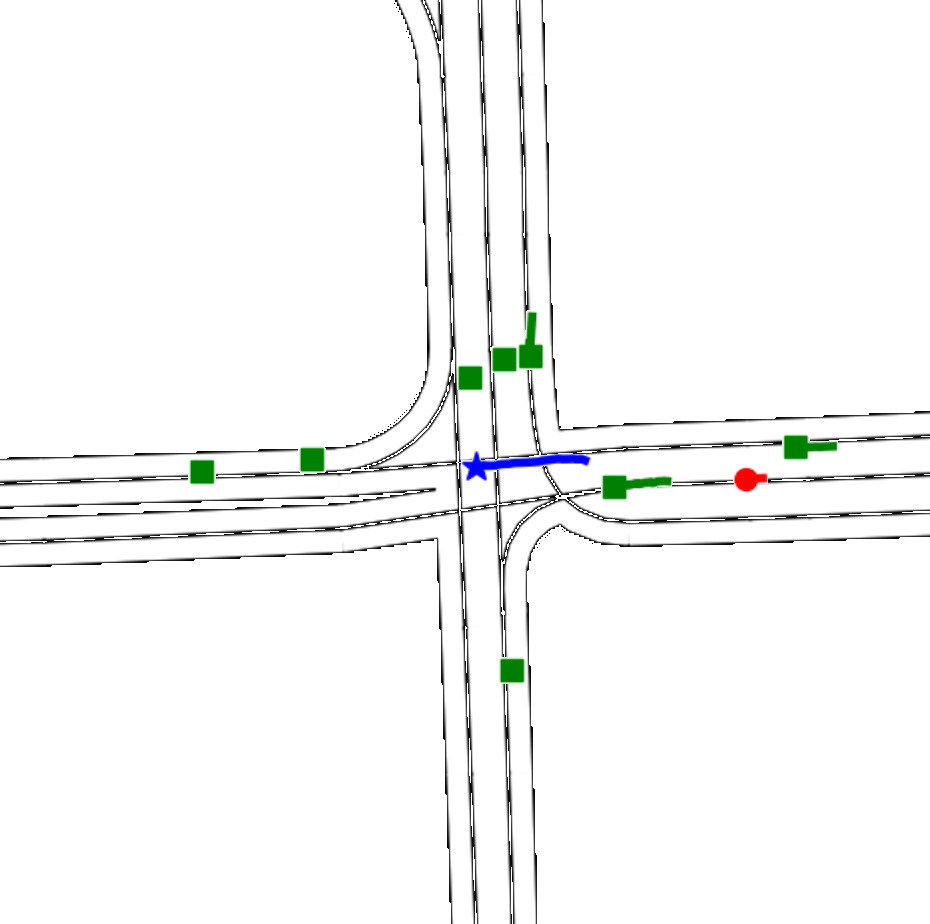}} &
    \fbox{\includegraphics[width=0.23\linewidth]{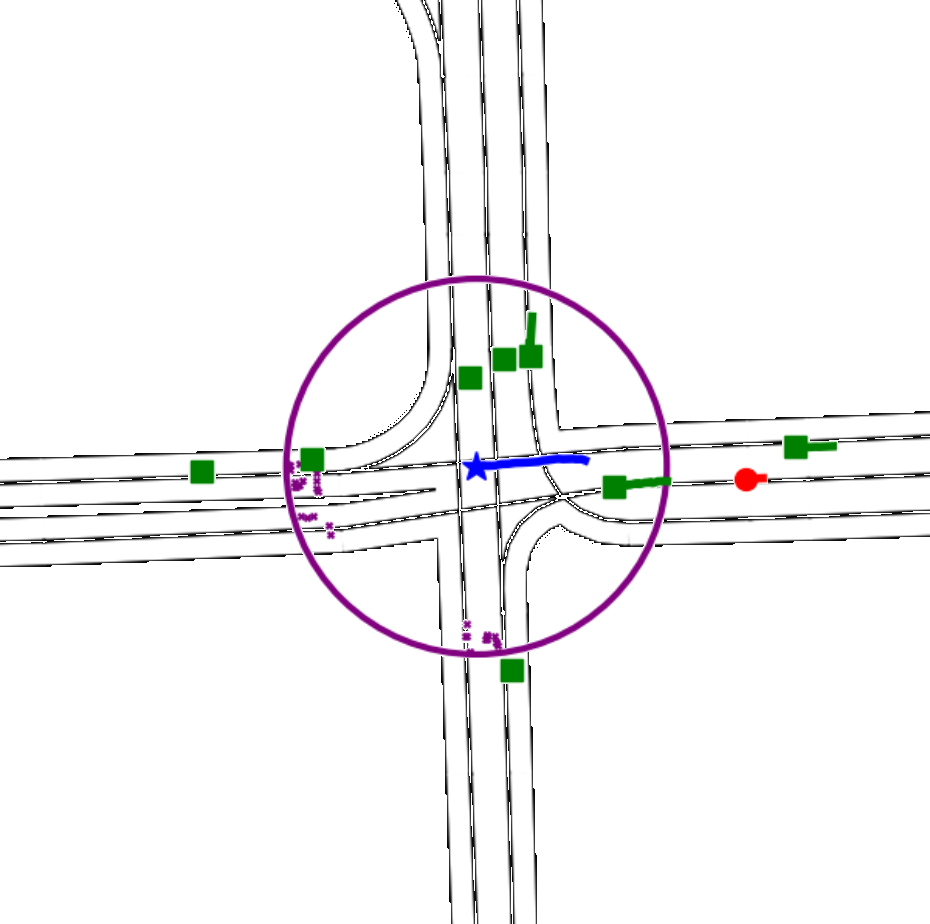}} &
    \fbox{\includegraphics[width=0.23\linewidth]{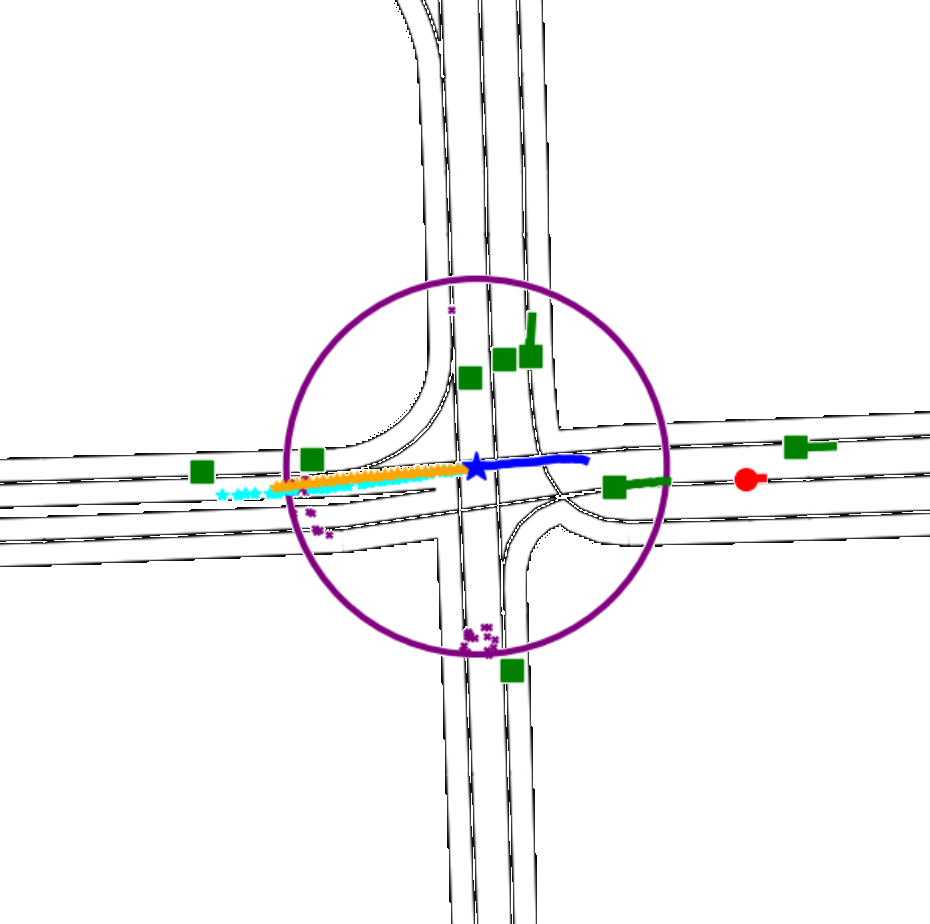}} & 
    \fbox{\includegraphics[width=0.23\linewidth]{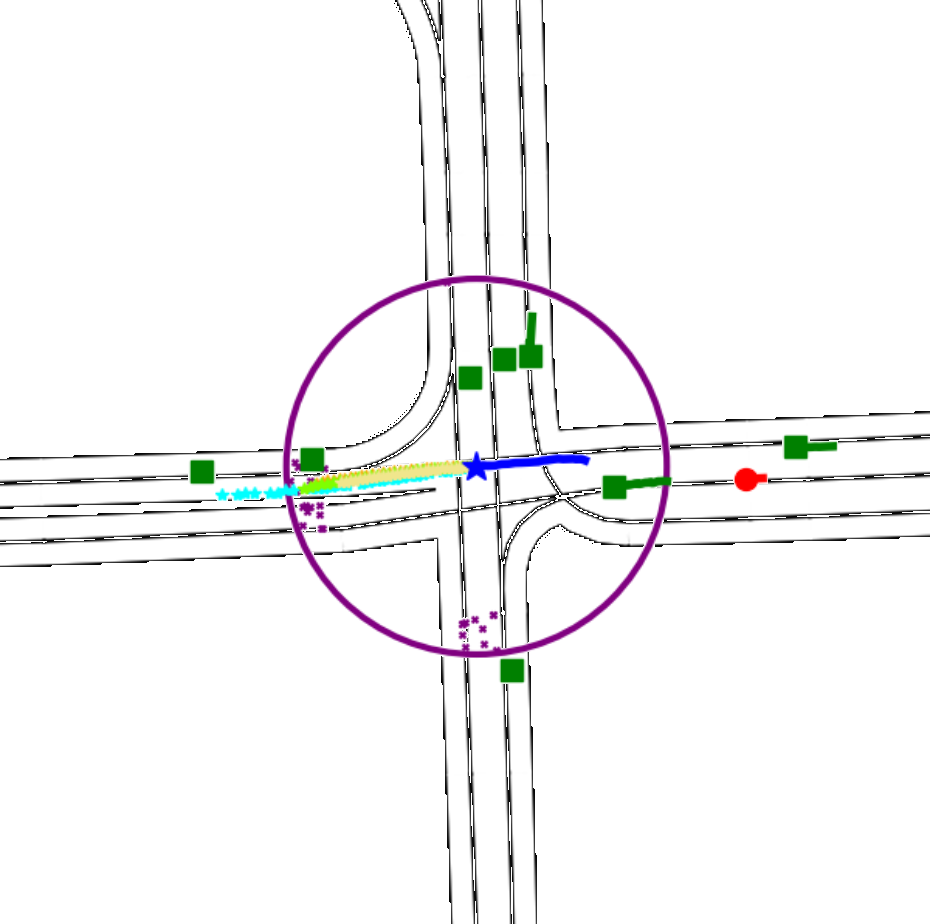}}
    \tabularnewline
    \fbox{\includegraphics[width=0.23\linewidth]{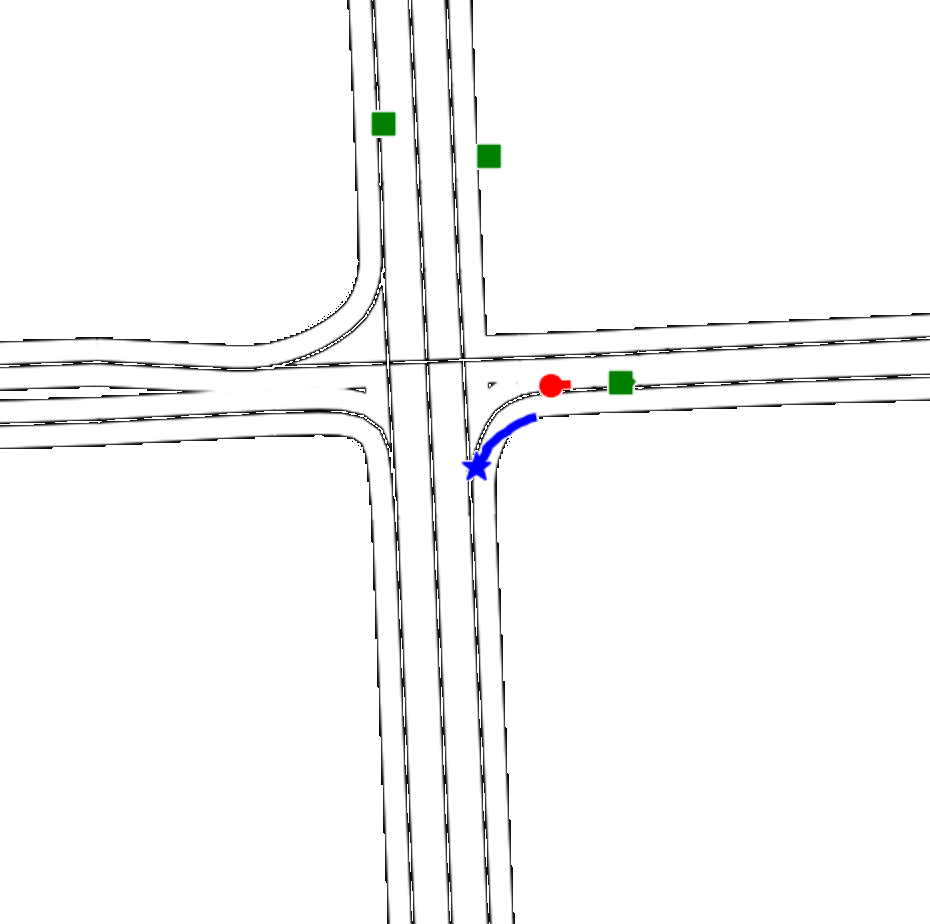}} &
    \fbox{\includegraphics[width=0.23\linewidth]{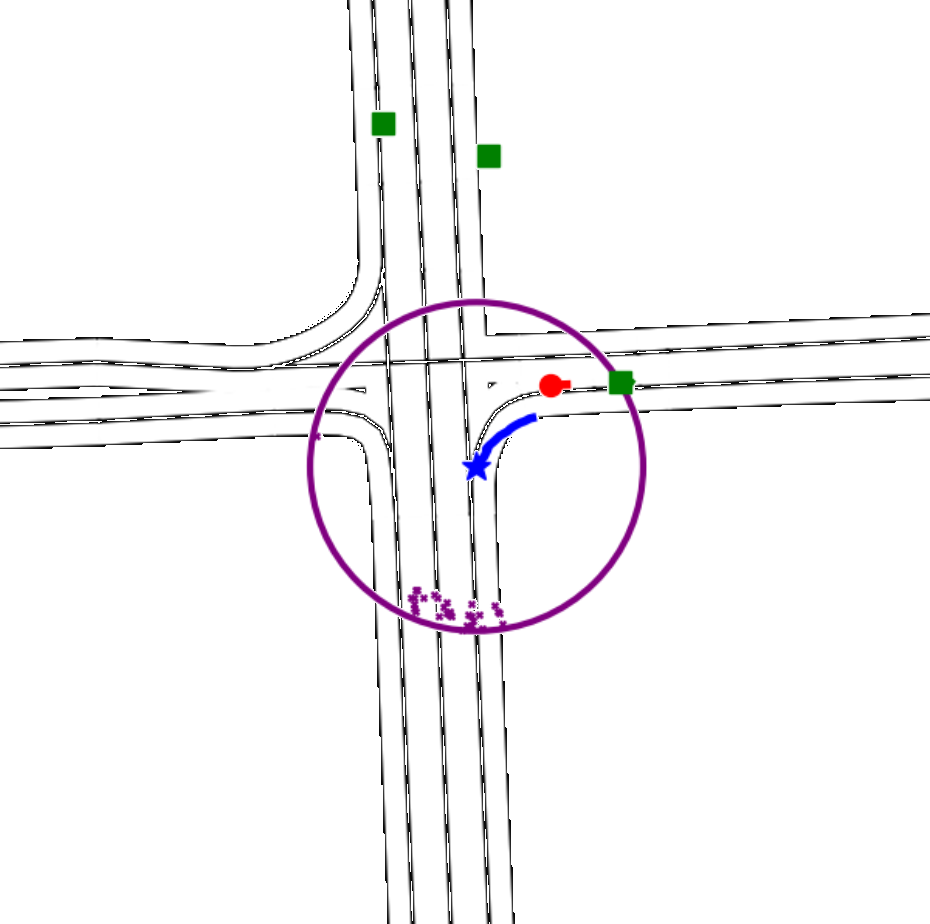}} &
    \fbox{\includegraphics[width=0.23\linewidth]{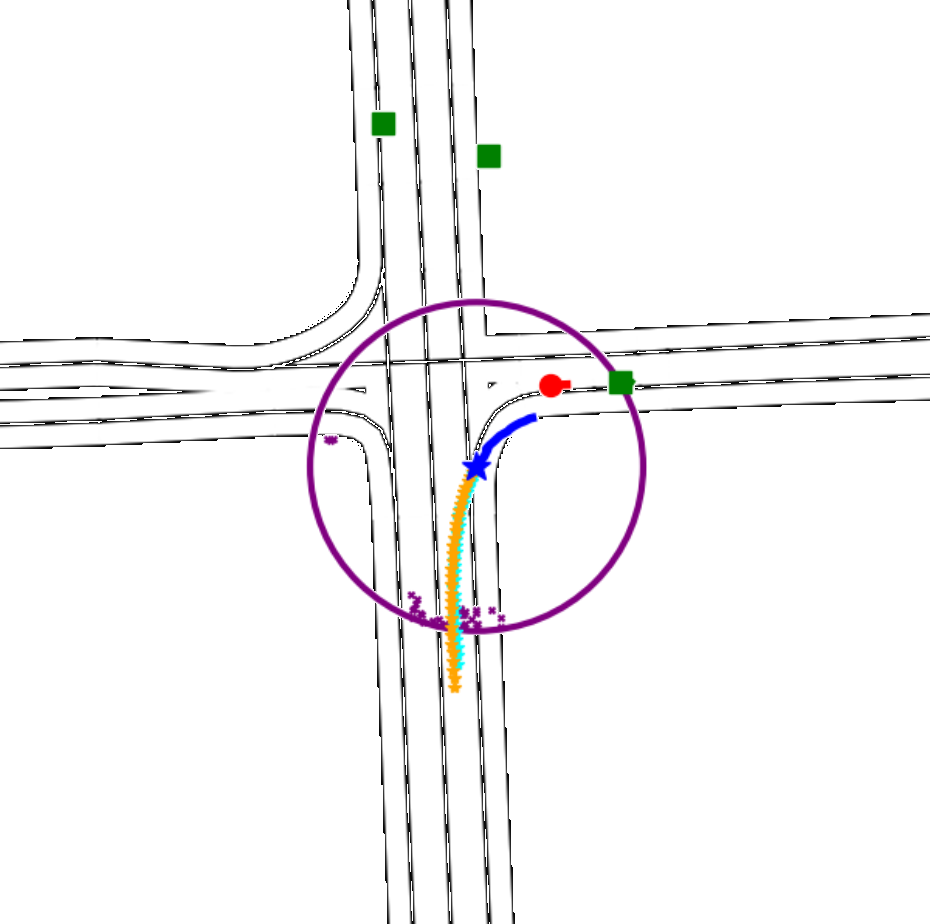}} & 
    \fbox{\includegraphics[width=0.23\linewidth]{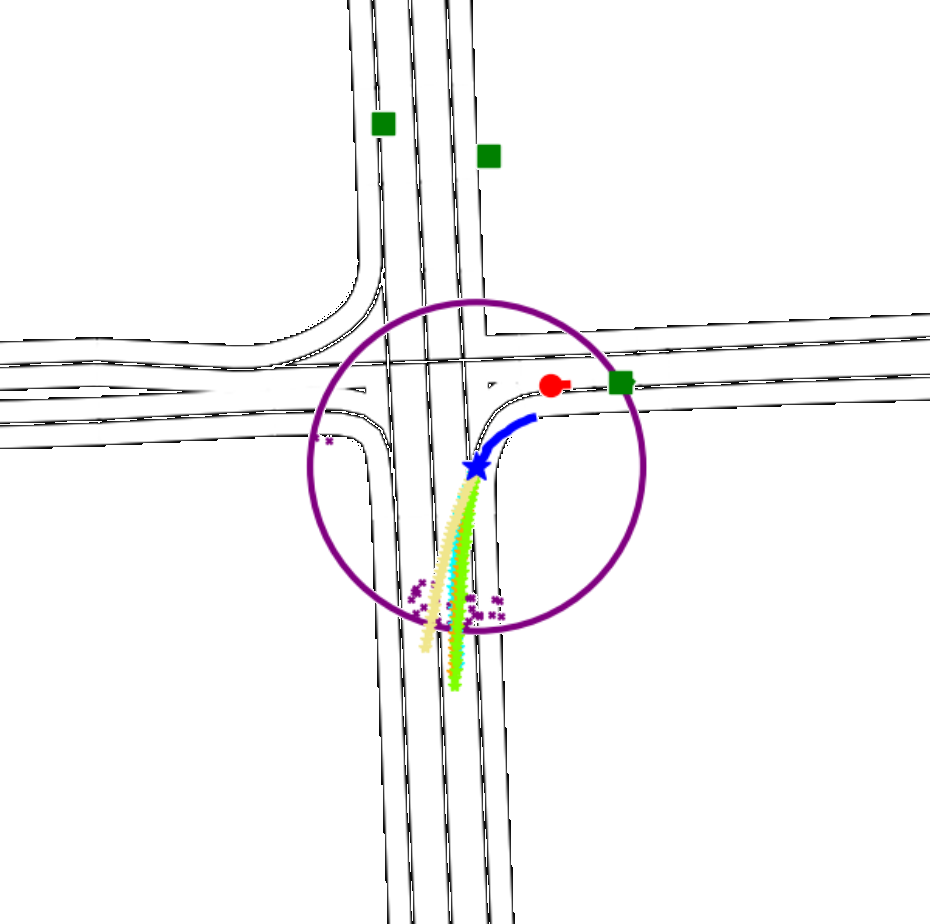}}
    \tabularnewline
    \fbox{\includegraphics[width=0.23\linewidth]{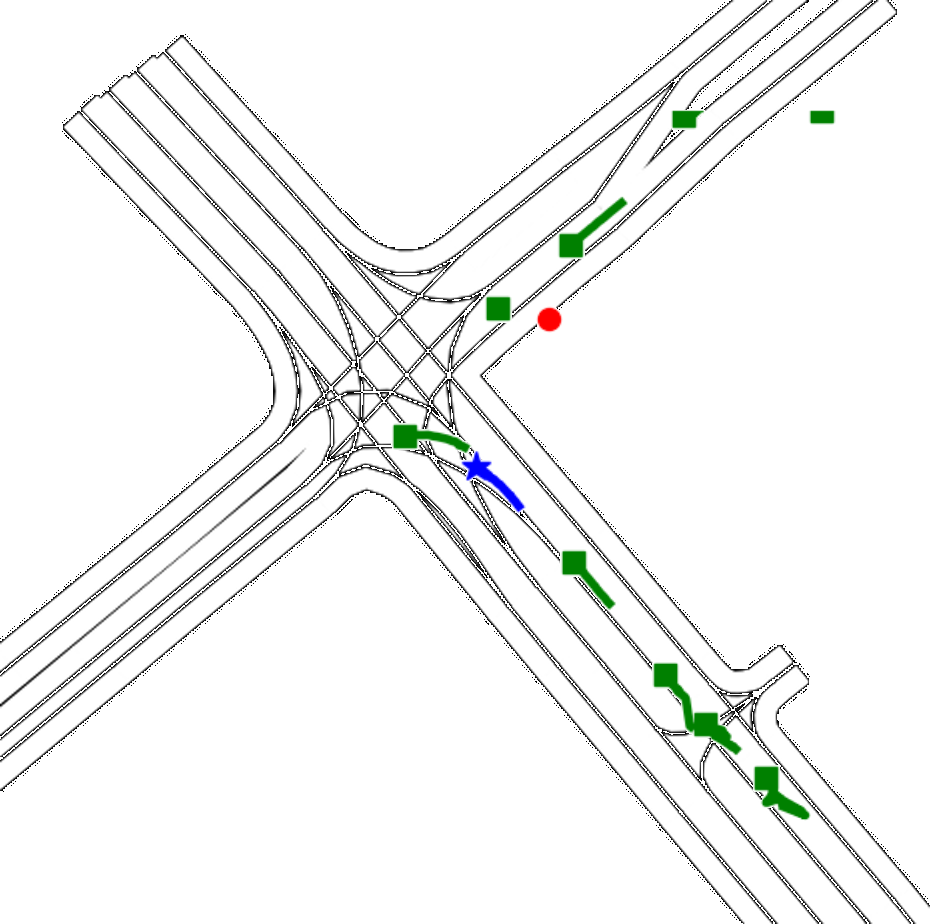}} &
    \fbox{\includegraphics[width=0.23\linewidth]{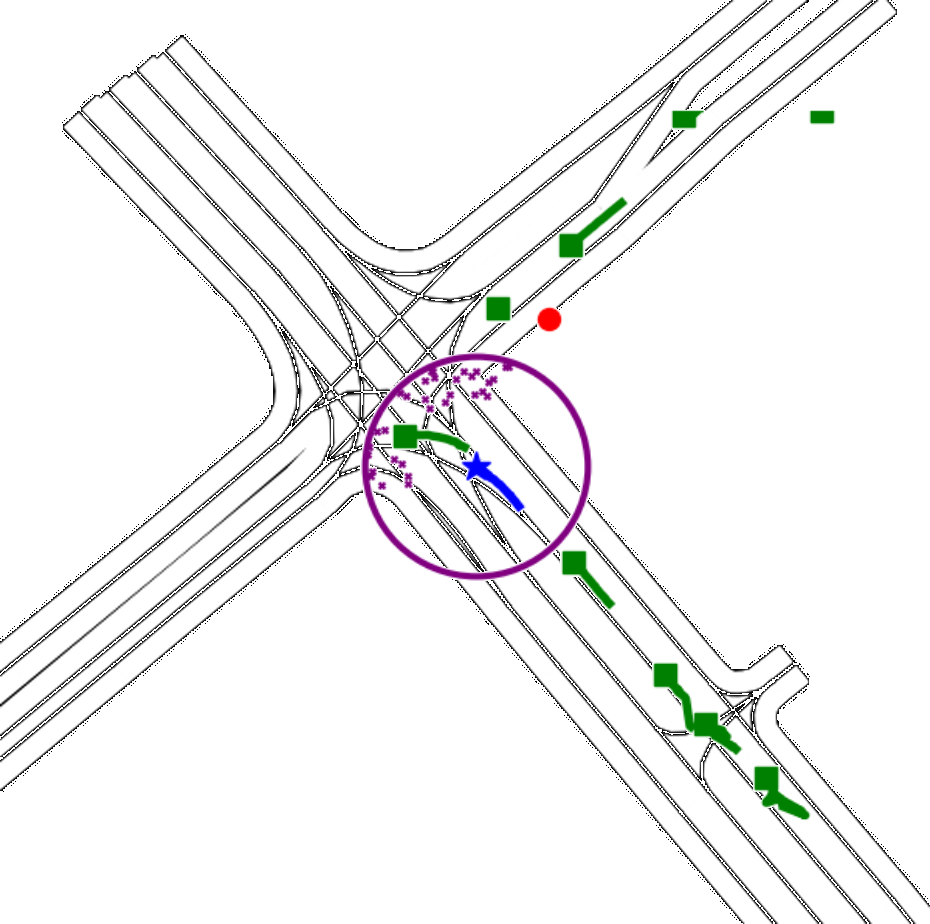}} &
    \fbox{\includegraphics[width=0.23\linewidth]{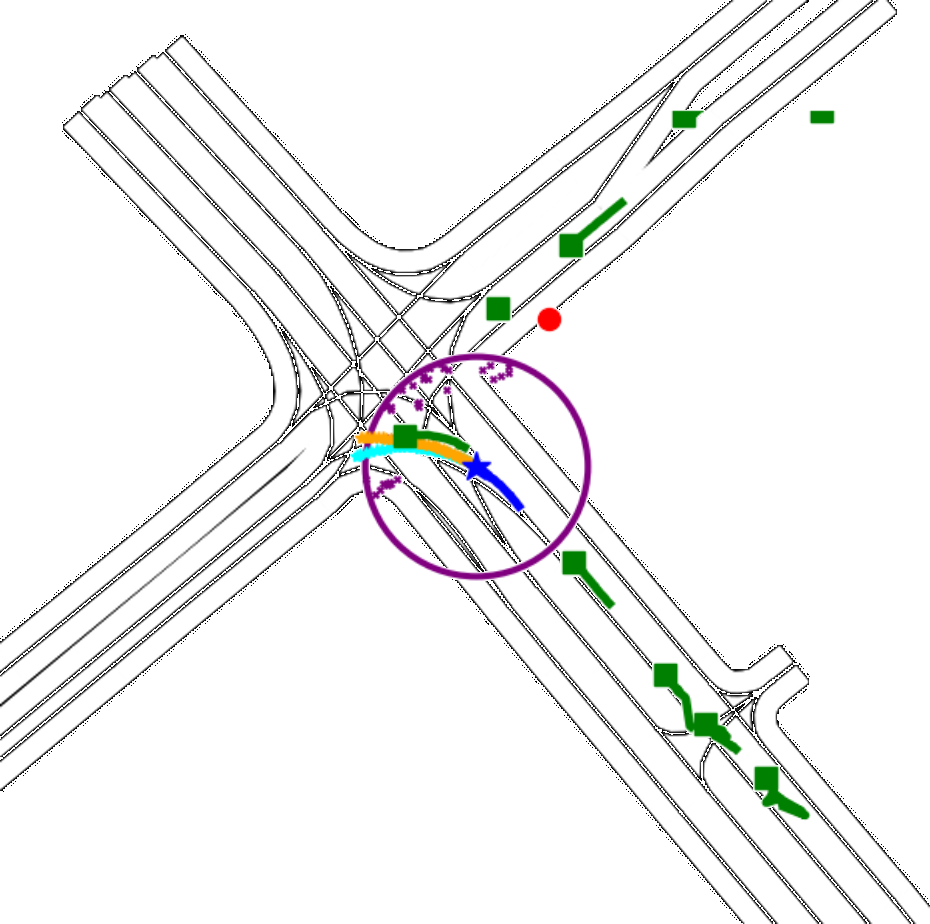}} & 
    \fbox{\includegraphics[width=0.23\linewidth]{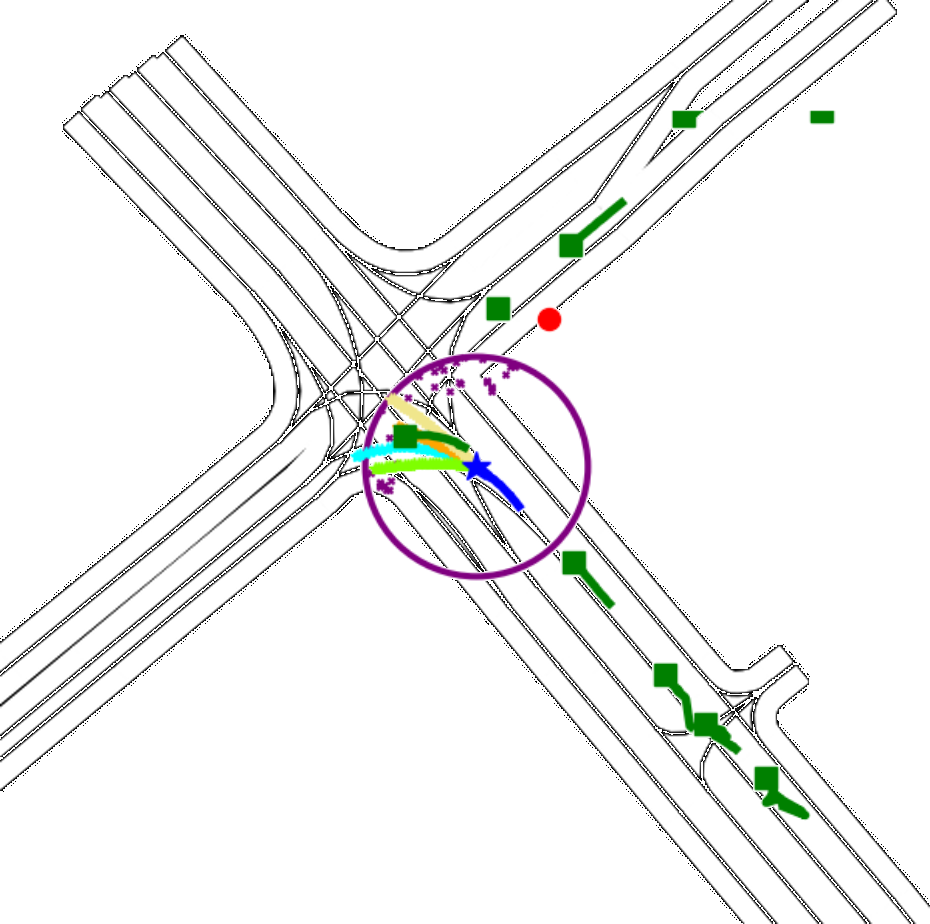}}
    \tabularnewline
    \fbox{\includegraphics[width=0.23\linewidth]{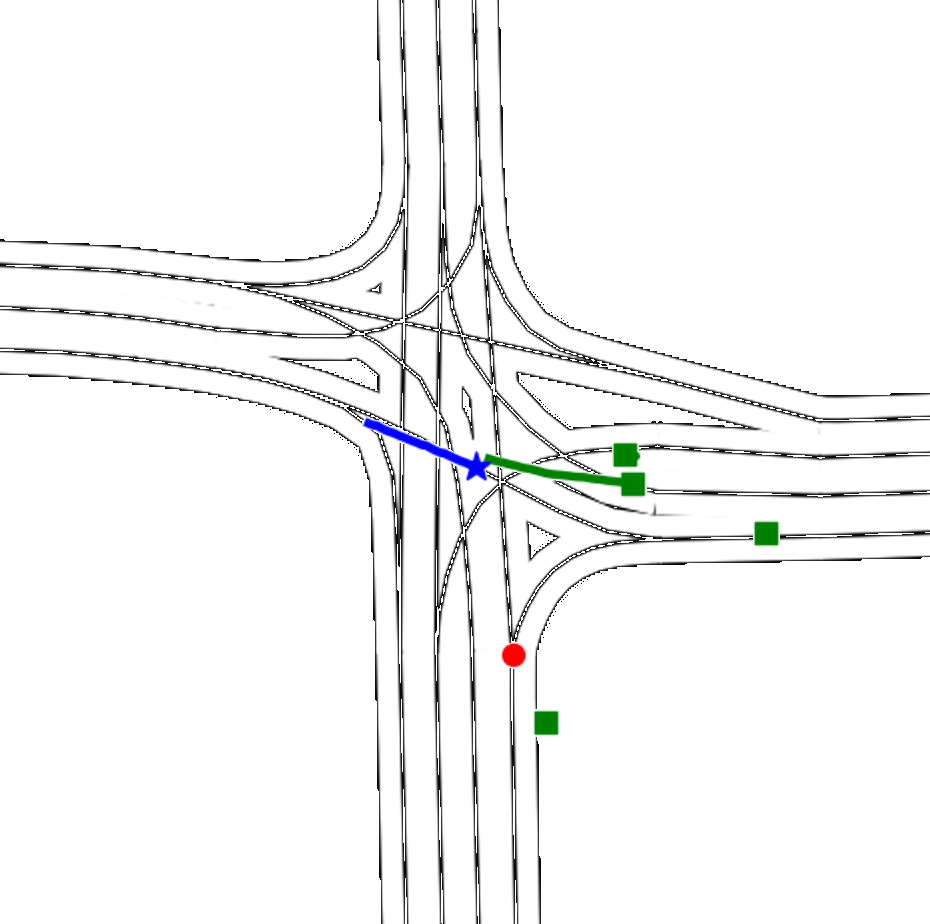}} &
    \fbox{\includegraphics[width=0.23\linewidth]{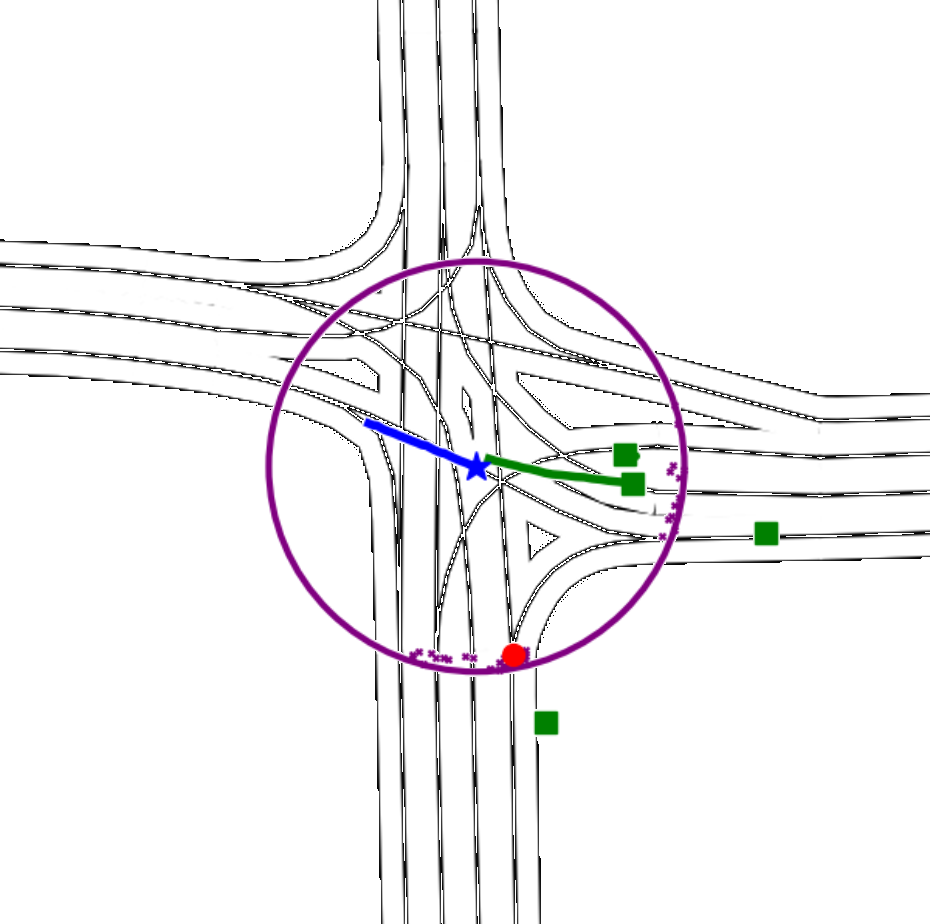}} &
    \fbox{\includegraphics[width=0.23\linewidth]{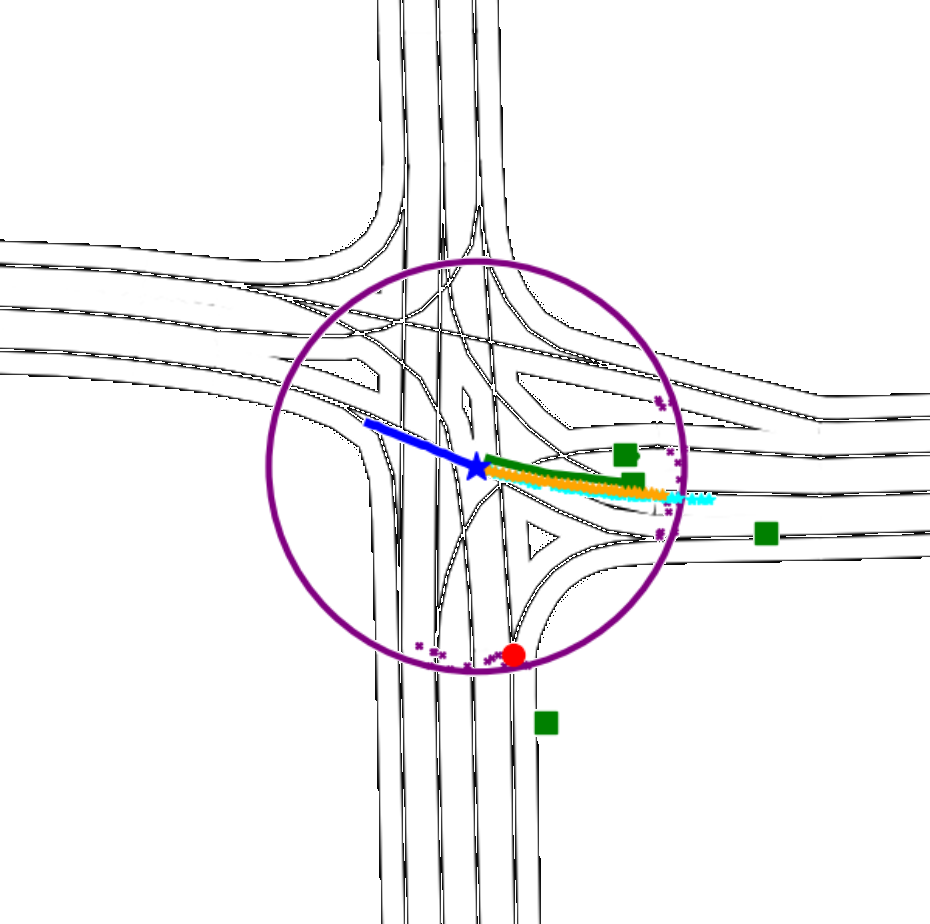}} & 
    \fbox{\includegraphics[width=0.23\linewidth]{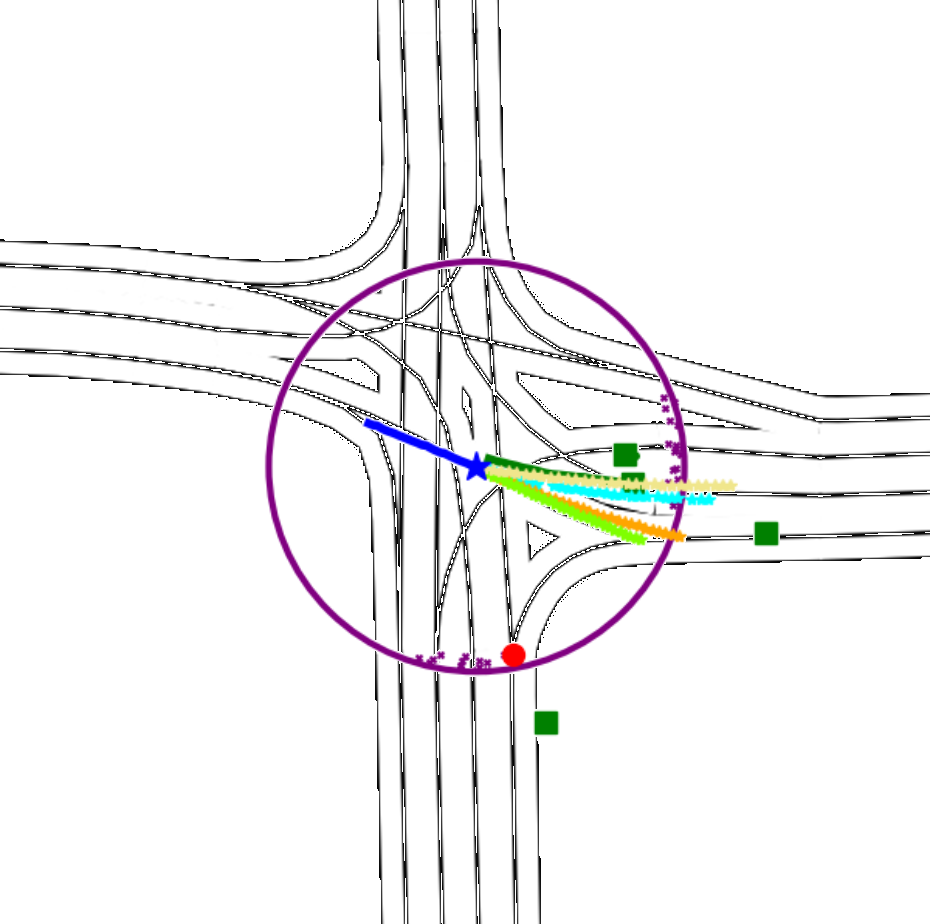}}
    \tabularnewline
    \tabularnewline
    Scene & Range and Goals & Uni-Modal Predictions & Multi-Modal Predictions (top-3) \tabularnewline
    \end{tabular}
    \caption{Qualitative Results on challenging scenarios using our best model. Qualitative Results using our best model. Legend: Our vehicle (\textcolor{red}{ego}), the \textcolor{blue}{agent} we are predicting its future, and \textcolor{ForestGreen}{other agents}. The \textcolor{cyan}{real} trajectory, the \textcolor{orange}{prediction}, the estimated action \textcolor{purple}{range} and potential \textcolor{purple}{goal-points}. Markers symbolize current positions.
    As we can see the predicted Goal points serve as a good guidance to our model, which can predict reasonable trajectories in presence of multiple agents and challenging scenarios. We show the top-3 from the 6 multi-modal predictions based on their confidence.}
    \label{fig:results}
\end{figure*}

As shown in in Table~\ref{tab:ablation}, our models achieve competitive results, and the proposed use of minimal HD map information (feasible area and goal points), leads to a better performance; we believe this is due to the incorporation of such prior knowledge and features about the target agent. We prove this qualitatively in App.~Fig.~\ref{fig:results} and Fig.~\ref{fig:results_teaser}, where we can see how the estimated goal points represent a good guidance for the model.\\

\textbf{Efficiency discussion} To the best of our knowledge, very little methods report efficiency-related information~\cite{gilles2021gohome, gilles2021home, liu2021mmtransf, gao2020vectornet}, moreover, comparing runtimes is difficult, as only a few competitive methods provide code and models. The Argoverse Benchmark~\cite{argobench} provides insightful metrics about the model's performance, mainly related with the predictions error, however, there are no metrics about efficiency (i.e. model complexity in terms of parameter or FLOPs, inference time per scene). In the self-driving context, we consider these metrics as important as the error evaluation because, in order to design a reliable Autonomous Driving system, we must produce \textbf{reliable predictions on time}, meaning the inference time (related to model's complexity and inputs) is crucial. SOTA methods already provide predictions with an error lesser than 1 meter in the multi-modal case. In our opinion, an accident will rarely happen because some predictions are offset by one or half a meter, but rather because lack of coverage or delayed response time.

Similar to image classification, where model \textbf{efficiency} depends on its accuracy and parameters/FLOPs, we use the same criteria to compare models. We show the efficiency comparison with other relevant methods in Table~\ref{tab:effcomp}. We calculate FLOPs using the relation: $\text{GMACs} \approx 0.5 * \text{GFLOPs}$ using \url{https://github.com/facebookresearch/fvcore}. Some minor operations were not supported, yet, their contributions to the number of FLOPs were residual (i.e. LeakyReLu activations) and ignored.

As we show, and was introduced by Gao~et al.~\cite{gao2020vectornet}, CNN-based models for processing the HD map information are computationally complex (i.e. the different ResNet18~\cite{he2016deep} networks with kernel size 3 and map resolution 100/400). 
We do not require CNNs to extract features from the HD map since we use our map-based feature extractor to obtain the feasible area and goal points. Moreover, these features are interpretable in comparison with CNNs high-dimensional outputs.
Note that for graph-based models like GOHOME~\cite{gilles2021gohome} and VectorNet~\cite{gao2020vectornet}, the FLOPs depends on the number of vector nodes or agents, and polylines/lanelets in the scene. In the case of VectorNet~\cite{gao2020vectornet}, the FLOPs increase linearly with the number of predicting targets as $0.041G \times n$ (in this case $n=1$).
Our model requires less computation than VectorNet~\cite{gao2020vectornet}, uses minimal HDMap information and achieves essentially the same performance. 

However, we must mention the \textbf{computational limiations} of our approach: our LSTM model is non-parallelizable, therefore it might not be suitable for real time applications.
The (Set) Transformer is much faster than RNN/LSTM/GRU (in a sense of wall clock time, especially when the number of agents $n$ is smaller than the dimensionality of the vector representations $d$) and produces better results~\cite{vaswani2017attention,lee2019set}.
In both variants, LSTM and Set Transformer, self-attention is used with a dynamic number of input agents, this typically implies a quadratic growth in complexity with the number of agents in the scenes~\cite{vaswani2017attention}, yet, this only applies to the MHSA layers. Moreover, Set Transformer~\cite{lee2019set, girgis2021autobots} reduces the computation time of self-attention from quadratic to linear in the number of elements in the set.

We provide \textbf{qualitative} results in Fig.~\ref{fig:results_teaser} and Fig.~\ref{fig:results}, showing challenging scenarios with multiple agents and complex topology, and different modality predictions, we also show the agent's predicted range of action.

\section{Conclusions}
\label{sec:conclusion}

Using HD map and trajectories information leads to SOTA results on the most relevant self-driving motion prediction benchmarks, however, these do not consider efficiency (i.e. inference time per scene, FLOPs). In this work, we achieve competitive results on Argoverse 1.0 Benchmark using compact attention-based models, which are more efficient in terms of FLOPs than previous approaches. We use map-based feature extraction to improve performance by feeding interpretable prior-knowledge into the model without increasing model complexity substantially and using these as an alternative to black-box CNN-based map processing.

In future work, we will improve our map-based features to obtain more interpretable priors (goal points, lane intersections, dynamic map elements as light signals) and regularization considering the feasible area to feed into the model as well as an enhanced version of the attention mechanisms and multimodal decoder to improve further the performance. We hope that our ideas can serve as a reference for designing efficient and interpretable MP methods. 

\clearpage

\section*{Acknowledgment}

This work has been funded in part from the Spanish MICINN/FEDER through the Techs4AgeCar project (RTI2018-099263-B-C21) and from the RoboCity2030-DIH-CM project (P2018/NMT- 4331), funded by Programas de actividades I+D (CAM), cofunded by EU Structural Funds and Scholarship for Introduction to Research activity by University of Alcalá.

Marcos Conde is with the Computer Vision Lab, Institute of Computer Science, University of Würzburg, Germany. Supported by the Humboldt Foundation.

{\small
\bibliographystyle{ieee_fullname}
\bibliography{bibliography.bib}
}


\end{document}